\documentclass{article} 
\usepackage[final]{colm2026_conference}

\usepackage{microtype}
\usepackage{hyperref}
\usepackage{url}
\usepackage{booktabs}

\usepackage{graphicx} 
\usepackage[table]{xcolor}
\usepackage{seqsplit}
\usepackage{wrapfig}
\usepackage{amsmath}

\usepackage{lineno}

\definecolor{darkblue}{rgb}{0, 0, 0.5}
\hypersetup{colorlinks=true, citecolor=darkblue, linkcolor=darkblue, urlcolor=darkblue}

\title{Round-trip Reinforcement Learning: Self-Consistent Training for Better Chemical LLMs}

\author{Lecheng Kong\thanks{Now at Amazon.} \\
Washington University in St.~Louis \\
\texttt{jerry.kong@wustl.edu} \\
\And
Xiyuan Wang \\
Peking University \\
\texttt{wangxiyuan@pku.edu.cn} \\
\And
Yixin Chen \\
Washington University in St.~Louis \\
\texttt{ychen25@wustl.edu} \\
\And
Muhan Zhang \\
Peking University \\
\texttt{muhan@pku.edu.cn}
}

\DeclareMathOperator*{\argmax}{arg\,max}

\begin{document}

\ifcolmsubmission
\linenumbers
\fi

\maketitle

\begin{abstract}
Large Language Models (LLMs) are emerging as versatile foundation models for computational chemistry, handling bidirectional tasks like reaction prediction and retrosynthesis. However, these models often lack round-trip consistency. For instance, a state-of-the-art chemical LLM may successfully caption a molecule, yet be unable to accurately reconstruct the original structure from its own generated text. This inconsistency suggests that models are learning unidirectional memorization rather than flexible mastery. Indeed, recent work has demonstrated a strong correlation between a model's round-trip consistency and its performance on the primary tasks. This strong correlation reframes consistency into a direct target for model improvement. We therefore introduce Round-Trip Reinforcement Learning (RTRL), a novel framework that trains a model to improve its consistency by using the success of a round-trip transformation as a reward signal. We further propose an iterative variant where forward and reverse mappings alternately train each other in a self-improvement loop, a process that is highly data-efficient and notably effective with the massive amount of unlabelled data common in chemistry. Experiments demonstrate that RTRL significantly \textbf{boosts performance and consistency} over strong baselines across supervised, self-supervised, and synthetic data regimes. This work shows that round-trip consistency is not just a desirable property but a trainable objective, offering a new path toward more robust and reliable foundation models.
\end{abstract}

\section{Introduction}
\label{sec:intro}
Large Language Models (LLMs) are emerging as a powerful class of versatile and generalizable foundation models for computational chemistry~\citep{fangmol,cao2025instructmol,wang2025survey}. A key advantage is their ability to provide a more flexible and intuitive interface for scientists, enabling interaction with complex data through natural language. This is powered by their unique capability to process and generate information across diverse chemical modalities, from structured SMILES strings to unstructured experimental procedures~\citep{zhao2025developing,pei2024biot5+}. By unifying these disparate tasks and data types within a single framework, LLMs promise to significantly accelerate the pace of scientific discovery.

A prominent theme enabled by this new generation of models is the modeling of \textit{bidirectional} chemical tasks, such as translating between textual descriptions and molecular structures or mapping between forward synthesis and retrosynthesis~\citep{pei2024enhanced,edwards-etal-2024-l,liu2024evaluating}. More than just a capability, however, this bidirectionality allows us to examine a model's depth of understanding through the crucial property of \textbf{round-trip consistency} (RTC). A model exhibits RTC if, for instance, after generating a textual caption for a given molecule, it can then take that caption as input and accurately reconstruct the original molecular structure. A failure to pass this test demonstrates a critical weakness: it relies on shallow, uni-directional memorization of statistical patterns rather than developing a flexible and abstract mastery of the underlying chemical principles. This distinction between memorization and true understanding provides critical insight into the strong positive correlation observed in recent work between a model's RTC and its overall task performance~\citep{liu2024evaluating,allamanis2024unsupervised}. A consistent model, by necessity, must learn the fundamental rules and relationships governing the data. For example, a model that truly understands an underlying reaction mechanism can reason bidirectionally, leading to both stronger round-trip consistency and superior generalization compared to a model that has only memorized common, unidirectional pairs. Ultimately, this deeper, more principled understanding is a prerequisite for scientific adoption, as high-stakes applications like therapeutic design demand models with exceptional overall knowledge utilization, reliability, and logical coherence.

While the importance of this property has been recognized in various domains~\citep{rennie2020unsupervised,somers-2005-round}, existing work has primarily leveraged it in two ways, one as a post-hoc evaluation metric~\citep{rennie2020unsupervised} and the other as a data augmentation technique~\citep{alberti2019synthetic}. Consequently, a method that can \textbf{algorithmically} enforce and \textbf{iteratively} improve round-trip consistency directly within the model's training loop has remained a significant, unaddressed challenge.

To address this gap, we introduce \textbf{Round-Trip Reinforcement Learning} (RTRL), a novel framework designed to directly optimize for round-trip consistency. Our approach frames the problem as a self-supervised task where a forward model is trained to produce outputs that a backward model can successfully map back to the original input. \textbf{The success of this reverse mapping serves as a reward signal to improve the forward model.} The core insight is that this process compels the model to build a deeper and more coherent internal representation of the chemical world. Rather than learning a shallow, unidirectional statistical mapping, the model must learn the underlying, bidirectional relationship that connects entities. For example, consider a case where the forward model generates an ambiguous output $B$ from an input $A$. The reverse model, attempting to reconstruct the input, may then produce $A'$ where $A'\neq A$. RTRL uses the resulting low reward to penalize the initial $A\rightarrow B$ generation. To maximize its reward, the forward model must learn to discard ambiguous outputs and instead produce a clearer, more concrete representation $B^*$ from which the original input $A$ is recoverable. We further propose an iterative variant of RTRL where the forward and backward functions \textbf{swap roles} to continuously solidify the model's knowledge in a self-improving paradigm.

A significant advantage of the RTRL framework is its adaptability to different data availability. Primarily, the training process only requires access to data from a single domain (e.g., a list of known molecular products), without needing labelled pairs (e.g., their corresponding reactants), to improve the ability to generate correct responses (e.g., enhancing retrosynthesis ability by accessing only the products). Furthermore, this paradigm facilitates strong improvement over the base model in cases where data are either \textit{supervised, self-supervised, or synthetic}. In our empirical evaluation, RTRL boosts self-consistency in terms of exact match by up to 52\%, improves primary task performance by up to 55\%. We also show the continuously improved efficacy of iterative LLM via self-play. These results validate our central thesis: that enforcing round-trip consistency is a powerful mechanism for unlocking its latent knowledge, leading to more robust and credible chemical foundation models.
\section{Preliminary}
\label{sec:pre}
\textbf{Molecule data and tasks:} In this paper, we primarily work with data from the chemical domain, apart from regular chemical text, the molecules are represeted in Simplified Molecular Input Line Entry System (SMILES) sequence, it provides 1-D unambiguous representation of a molecule, which is a good fit for LLM processing. SMILES can also represents a Reaction in the format of ``\textit{Reactants$>$Reagents$>$Products}'', where \textit{Reactants}, \textit{Reagents} and \textit{Products} are chemicals in SMILES separated by dots. Concrete examples of SMILES and Reaction SMILES can be found in Appendix~\ref{app:data}. Note that unlike a traditional reaction equation, where the LHS and RHS of the equation must have the same elements, Reaction SMILES can omit secondary products and reactants to focus on the core transformation, so the elements in reactants and products might not match exactly.

This paper focuses on two pairs of bidirectional tasks, molecule captioning (molecule to description) verus text-based molecule generation (description to molecule) and reaction prediction (reactants to products) versus retrosynthesis (products to reactants). Modern chemical language models usually are equipped with all four functions~\citep{fangmol,zhao2025developing,pei2024biot5+}, as jointly training them potentially results in positive knowledge transfer.

This paper uses Group Relative Policy Optimization (GRPO) as the RL method, which is described in details in Appendix~\ref{app:grpo}.
\section{Round-Trip Consistency is Critical in Bidirectional Systems}
\label{sec:rtcwhy}
\begin{wraptable}{r}{7cm}

\resizebox{\linewidth}{!}{
\begin{tabular}{lccc}
\toprule
Exact Match & Retro.$\rightarrow$Re. Pred. & Re. Pred.$\rightarrow$Retro. & Cap.$\rightarrow$Gen. \\
\midrule
ChemDFM  & 0.032  & 0.547 & 0.170 \\
LlaSMol  & 0.535  & 0.304 & 0.041 \\
\bottomrule
\end{tabular}}
\vspace{-10pt}
\caption{RTC of SOTA Chemical LLMs.}\label{tab:con_exp}
\end{wraptable} 
The core principle in this paper is round-trip consistency (RTC)~\citep{somers-2005-round,yung2025round}. \textit{This refers to a system's ability to take an input, map it to another domain, and then reliably reconstruct the original input from that output.} The ability to maintain this consistency is a crucial benchmark for a model's depth of understanding. Failures in this area often suggest that a model is relying on shallow memorization rather than developing a high-level understanding of the underlying chemical knowledge~\citep{alberti2019synthetic,hong2025consistencychecker}.

Recent work has demonstrated that a model's overall performance is often \textbf{highly correlated with its consistency}; models demonstrating round-trip consistency tend to be more accurate and reliable~\citep{liu2024evaluating,allamanis2024unsupervised,hong2025consistencychecker}. This raises a critical question: \textit{How consistent are current chemical LLMs?}
We present a case study on two SOTA chemical LLMs, ChemDFM~\citep{zhao2025developing} and LlaSMol~\citep{yullasmol}. We conducted three round-trip experiments where we start with a molecule (molecules), perform forward and then backward transformation using the LLM, and eventually test if the output matches the original input exactly. In Table~\ref{tab:con_exp}, we can see that even the most recent LLM with broad chemical knowledge struggles to output round-trip consistent results, revealing a significant consistency gap. Also, different models behave differently in RTC. This empirical evidence confirms that the consistency gap is not a minor issue but a systematic limitation in current models. Meanwhile, this also highlights a major opportunity: directly addressing this fundamental weakness could be a key to elevating the capabilities of chemical LLMs to the next level. And in this paper, we will introduce Round-trip Reinforcement Learning (RTRL) to do just that.
\section{Round-Trip Reinforcement Learning}
\label{sec:method}
In a supervised learning setting, the objective is to learn a mapping from an input domain $\mathcal{X}$ to an output domain $\mathcal{Y}$. Given a dataset of paired examples $(x_i, y_i)$, the goal is to learn the parameters $\theta$ of a model $f:\mathcal{X}\rightarrow\mathcal{Y}$ that minimizes a predefined loss function, $\mathcal{L}$. The objective is typically expressed as:
\begin{equation}
    \theta^*=arg\min_{\theta} \mathbb{E}_{(x,y)\sim\mathcal{D}}[\mathcal{L}(f_\theta(x),y)]\quad x_i\in\mathcal{X},y_i\in\mathcal{Y}
\end{equation}
In contrast to supervised learning, the principle of RTC provides a powerful \textbf{self-supervised objective}. This requires a system with bidirectional ability with a forward function $f$ and a corresponding backward function $g:\mathcal{Y}\rightarrow\mathcal{X}$. Then, a round-trip transformation can be defined as:
\begin{equation}
    \text{Forward:} y=f(x), \text{Backward:} x'=g(y)=g(f(x))
\end{equation}
The system is considered consistent if the final output $x'$ is highly similar to the original input $x$. Therefore, the learning goal is to optimize the functions $f$ and $g$ to maximize their similarity. This can be formulated as maximizing the expectation of a similarity function $s(\cdot,\cdot)$:
\begin{equation}
    \max_{f,g}\mathbb{E}_{x\sim\mathcal{X}}[s(x, g(f(x)))]
\end{equation}
A key advantage of this formulation is that the objective can be optimized using \textbf{only inputs from} $\mathcal{X}$, without requiring corresponding ground-truth labels from $\mathcal{Y}$. This naturally facilitates self-supervised learning, and we will discuss this further in Appendix~\ref{app:rtrl_usecase}.

This principle is particularly well-suited for Large Language Models (LLMs), where a single model can perform diverse functions conditioned on different prompts. The forward and backward functions, $f$ and $g$, are not separate models but are instantiations of the same LLM, parameterized by $\theta$, conditioned on a pair of inverse prompts, $t_f$ and $t_g$, and we can define 
\begin{equation}
    f(x)\equiv LLM_\theta(x,t_f),\quad g(y)\equiv LLM_\theta(y,t_g)
\end{equation}
and we can rewrite the task to maximize RTC as,
\begin{equation}\label{eq:opt}
    \theta^*=\argmax_{\theta} \mathbb{E}_{x\sim\mathcal{X}}[s(x, LLM_\theta(LLM_\theta(x,t_f),t_g))]
\end{equation}
Optimizing this objective refines the model's internal parameters to ensure that the knowledge used for the forward and reverse tasks is coherent. Such a process encourages a shift from unidirectional memorization toward a more robust, bidirectional understanding.

\begin{figure*}[t]
    \centering
    \includegraphics[width=\textwidth,trim={0.5cm 9cm 0.5cm 1.3cm},clip]{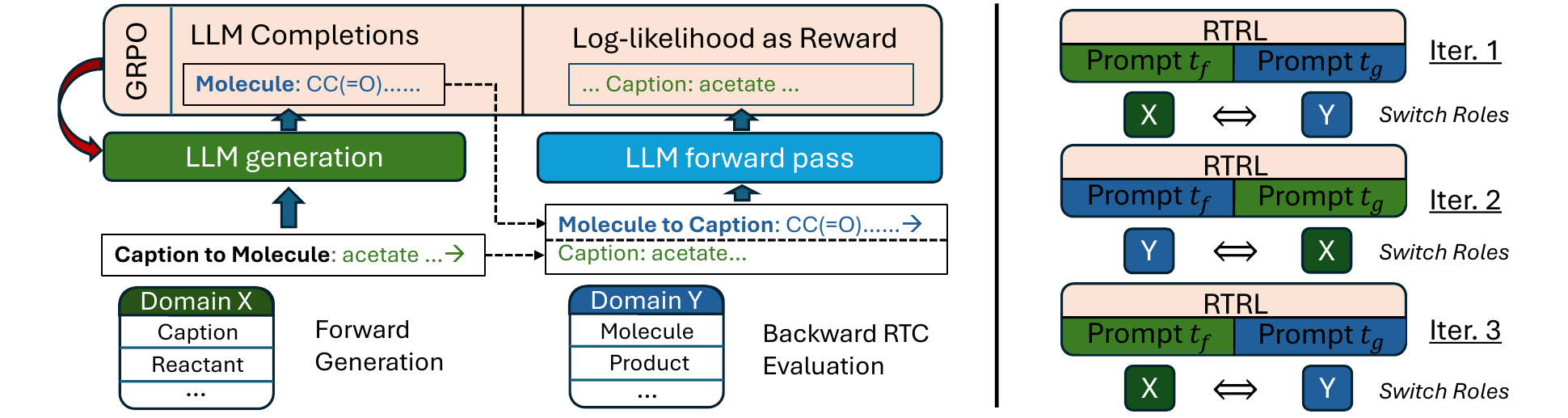}
    \caption{Left: Round-Trip Reinforcement Learning Pipeline. An input $x$ is mapped to an output $y$ by a forward prompt. We then compute the generation likelihood of $x$ given $y$ and a backward prompt. The likelihood is used as the reward for RL. Right: In iterative RTRL, we switch the forward and back prompts as well as the input and output domains to achieve mutual improvement.}
    \label{fig:rtrl}
    \vspace{-10pt}
\end{figure*}
\subsection{Improving Round-Trip Consistency With Reinforcement Learning}
Optimizing Equation~\ref{eq:opt} is not straightforward, as the output of $LLM_\theta$ is a sampled text response, precluding the use of standard gradient-based methods. Hence, we propose to use Reinforcement Learning (RL) to train the system. While our framework is compatible with various policy gradient algorithms, we adopt Group-Relative Policy Gradient (GRPO) for its effectiveness in LLM training. The overall method pipeline is illustrated in Figure~\ref{fig:rtrl}.

A well-defined reward function is critical for any RL system. An intuitive and natural choice is directly using the objective proposed in Equation~\ref{eq:opt}. However, such a choice presents two significant challenges. First, it requires engineering a task-specific similarity function for the text sequence space, which may not capture true chemical similarity (e.g., the BLEU score is a poor metric to assess chemical similarity between SMILES). Second, it is computationally prohibitive, requiring two full autoregressive generation steps (forward and backward generation) for a single reward calculation. To handle the two problems, we propose a surrogate objective,
\begin{equation}
    \theta^*=\argmax_{\theta} p_\theta(x|LLM_\theta(x,t_f),t_g),
\end{equation}
where $p_\theta(x|y)$ is the LLM's conditional probability of generating $x$, given prefix $y$. Instead of generating a full backward response, we use the conditional likelihood of the original input $x$ as the reward. This formulation handles the two problems mentioned above directly. Now, it uses the LLM itself as the judge to evaluate the similarity. With a higher likelihood to generate $x$, it increases the overall likelihood that the output is similar to the original input. Such an approach does not require manually setting up a similarity function, avoiding human effort and flawed similarity design. This is inspired by recent work in LLM-as-a-judge~\citep{gu2025surveyllmasajudge}. Secondly, in this approach, the likelihood can be computed via a single forward pass, largely improving the overall efficiency.

Following this formulation, we use GRPO to optimize the objective. Specifically, we begin with two copies of the same LLM for the forward function $LLM_\theta$ and the backward function $LLM_\phi$, $\theta=\phi$ initially. The forward function is used as the policy model and will be updated by the GRPO training process. The backward function computes the likelihood of the original output given the forward function output. For efficiency and to stabilize the reward landscape, we follow standard RLVR approach to only update the policy parameters $\theta$, while keeping $\phi$ fixed. So, the final objective for the RL training is,
\begin{equation}\label{eq:obj}
    \theta^*=\argmax_{\theta} p_\phi(x|LLM_\theta(x,t_f),t_g).
\end{equation}
Then, for a input $x$, we generate a group of output $Y$ using $LLM_\theta(x,t_f)$, and the reward for each generation is computed as,
\begin{equation}
    r(y) = \sum_{i=1}^L \log p_\phi(x_i|x_{<i},y,t_g),\quad \forall y\in Y,
\end{equation}
where $L$ is the length of the input, and we normalize the reward by $L$ in practice to balance the impact of different input lengths. Then, the group of rewards is passed to GRPO for model optimization.

Modern RL for LLMs enforces format guardrails for safe and effective training, such as requiring a boxed answer for mathematical reasoning tasks or passing syntax checks for software engineering tasks. Following GRPO, we adapt its format reward to RTRL to implement this guardrail. In RTRL, the total reward for an input $x$ and a generated output $y$ is defined as:
\begin{equation}
    r(y) = \sum_{i=1}^L \log p_\phi(x_i|x_{<i},y,t_g) + \alpha F(y),
\end{equation}
where $F(\cdot)$ is a binary format reward function ($F(y)=1$ if the format is correct, and $0$ otherwise). In the context of our chemical tasks, for instance, this can be RDKit functionalities to validate and parse the output. Following the principles of GRPO, the format score must heavily outweigh the continuous likelihood reward to ensure strict adherence to these structural constraints. Because the round-trip log-likelihood is upper-bounded by 0 but lacks a strict lower bound, we systematically determine the scaling factor $\alpha$. Specifically, we compute the round-trip log-likelihood over a preliminary set of training steps to identify its empirical minimum value, $m$, and set $\alpha = -m$. This guarantees that a structurally invalid output, regardless of its round-trip likelihood, can never achieve a higher total reward than a format-correct output. More discussion on the format reward is in Appendix~\ref{app:format_reward}. We show that using format reward can easily avoid identity-mapping behavior for RTRL.

\subsection{Iterative RTRL for Mutual Improvement}
The framework described thus far details how Round-Trip Reinforcement Learning (RTRL) can enhance the consistency of a forward mapping, leveraging the dataset only from the input domain $\mathcal{X}$. This is a critical application in computational chemistry, where large databases of molecules, $X$, often exist without corresponding ground-truth labels, $Y$ (descriptions, synthetic pathways, etc.).

A natural extension of this framework is to consider a symbiotic relationship: \textit{can the now-improved forward model be used to enhance its inverse counterpart?} Once the forward generator $LLM(\cdot,t_f)$ has been refined, its ability to produce high-quality outputs makes it a \textbf{more reliable judge} for the reverse task. A more accurate judge provides a more meaningful reward signal, which in turn facilitates a more effective training process for the reverse mapping, $g:\mathcal{Y}\rightarrow\mathcal{X}$. This creates a virtuous cycle where each model bootstraps the other.

To implement this reverse training, a dataset from the target domain, $Y\subset\mathcal{Y}$, is required. Critically, this dataset Y does not need to be paired with the original dataset $X$. The two can be independent collections of data, potentially covering different subdomains of the problem space. This flexibility allows our framework to operate in various data regimes, see Appendix~\ref{app:rtrl_usecase} for a detailed discussion.

The iterative training procedure alternates between forward and reverse optimization phases. Starting with a model with parameters $\theta_k$ at iteration $k$, we first perform the standard RTRL, optimizing the objective in Equation~\ref{eq:obj}. In the next phase, the roles are reversed, and we optimize:
\begin{equation}
    \theta_{k+1}=\argmax_{\theta} p_{\phi_{k}}(y|LLM_\theta(y,t_g),t_f).
\end{equation}
The prompts $t_f$ and $t_g$ are \textbf{switched}, and $\phi_{k}$ and $\theta$ are initialized by $\theta_k$. We can then repeat this process to progressively refine both mappings to be more accurate and mutually consistent.

The RTRL framework is highly adaptable to various data availability scenarios. We briefly summarize its four primary operating regimes below and empirically validate in Section~\ref{sec:exp}, with a detailed discussion of its broader applications provided in Appendix~\ref{app:rtrl_usecase}:

\textbf{Single Mapping Enhancement:} Improves unidirectional tasks ($f: \mathcal{X} \rightarrow \mathcal{Y}$) using only unlabelled input data.
\textbf{Iterative Co-evolution:} Leverages unpaired, independent datasets from both domains, allowing the forward and reverse models to alternate roles and mutually improve.
\textbf{Supervised Regularization:} Acts as a structural regularizer when fully paired data $(X, Y)$ is available.
\textbf{Synthetic Self-Play:} Bootstraps entirely from an unlabelled seed dataset, using the model's own exploratory generations to autonomously search the problem space.
\section{Related Work}
\textbf{Chemical Language Model:} Earlier efforts such as MolT5~\citep{cao2025instructmol} and BioT5~\citep{pei2024biot5+} established language-to-molecule translation task. The surge of LLM leads to extensive efforts to adopt LLM to molecular learning and science discovery for its exceptional sequential modeling ability and generalizability~\citep{wang2025survey,yang2024molecule,cheng2021molecular}. Most works focus on collecting high-quality pre-training{\color{blue}~\citep{zhao2025developing,dey-etal-2025-mathtt,ye2025drugassist,xia2025naturelanguagemodeldeciphering, liu2024reactxtunderstandingmolecularreactionship,liu2024moleculargptopenlargelanguage} }and instruction-tuning~\citep{cao2025instructmol,fangmol,yullasmol,zhang2024chemllm} datasets to enhance models' ability to comprehend both chemical entities (molecule, SMILES, IUPAC names) and content (chemical bio-medical texts). Apart from the data perspective, recent works align LLM with structural or chemical knowledge to enhance native LLM's performance on chemical tasks~\citep{li2024molreflect,lin2024property,guo2025unimotunifiedmoleculetextlanguage}, and some work uses structural model like GNN to enhance the LLM~\citep{reactxt}. Concurrently, several works propose to use reinforcement learning or preference optimization to chemically align properties better~\citep{jang2024can,calanzone2025mol,gkoumas2024almol,lee2025mol}. Compared to existing works, we propose RTRL that only needs unlabelled data to significantly improve the performance, largely reducing the difficulty of training an effective chemical LLM.

\textbf{Round-trip Consistency.} Apart from the works introduced in Section~\ref{sec:rtcwhy}, enforcing this round-trip coherence shares conceptual roots with dual learning~\citep{duallearning} and cycle-consistency~\citep{cycle-con} in machine translation, including recent applications of round-trip GRPO for low-resource languages~\citep{rtgrpo}. Within the chemical domain, the concurrent work RTMol~\citep{rtmol} also explores a round-trip training mechanism. However, RTRL differs fundamentally in both computational efficiency and generality. RTMol calculates rewards using Tanimoto similarity, which necessitates a computationally expensive full autoregressive generation step and confines the method strictly to molecule output. In contrast, RTRL formulates the reward as a round-trip log-likelihood computed via a single, highly efficient forward pass. Because this likelihood formulation is \emph{representation-agnostic}, it applies uniformly to any text-based output within chemistry---SMILES strings, reaction SMILES, or natural-language captions---without designing task-specific similarity metrics. We validate RTRL within computational chemistry; extending it to fundamentally different domains is a promising direction we discuss in Appendix~\ref{app:rtrl_usecase}. We further provide a direct head-to-head comparison to RTMol under a matched setup in Section~\ref{sec:exp} (Table~\ref{tab:rtmol}), where RTRL attains comparable or better accuracy at substantially lower training cost.

\textbf{RL with Verifiable Rewards.} RL with verifiable rewards (RLVR) improves LLM reasoning using programmatically computable signals---e.g., answer correctness in math~\citep{shao2024deepseekmath} or unit tests in code---as the reward. Recent work extends RLVR beyond domains with clean verifiers via self-verification~\citep{liu2025trustverifyselfverificationapproach}, verifier-free token-probability rewards~\citep{yu2025rlprextrapolatingrlvrgeneral}, and internal-feedback signals that boost but then degrade reasoning, underscoring the value of grounded rewards~\citep{zhang2025freelunchrethinkinginternal}. RTRL differs from both: unlike standard RLVR it needs no domain-specific verifier, and unlike purely internal feedback its reward is structurally grounded in input reconstruction via the backward model's round-trip log-likelihood---requiring no labels or hand-designed metrics. RTRL is also compatible with RLVR: when verifiable rewards exist, the round-trip reward fuses with them for complementary signal (Section~\ref{sec:exp}).

We defer more related work on self-improving and RL to Appendix~\ref{app:more_rwork}.

\section{Experiment}
\label{sec:exp}
\begin{table*}[t]
\centering

\resizebox{1.0\textwidth}{!}{
\begin{tabular}{lcccccccc}
\toprule
Task / Model & BLEU $\uparrow$ & Lev. $\downarrow$ & Exact Match $\uparrow$ & MACCS SIM. $\uparrow$ & RDKit SIM. $\uparrow$ & Morgan SIM. $\uparrow$ & FCD $\downarrow$ & Validity $\uparrow$ \\
\midrule
\rowcolor{gray!25} \multicolumn{9}{l}{\textit{USPTO-Mixed Retrosynthesis$\rightarrow$Reaction Prediction}} \\
  ChemDFM & 0.535 & 59.355 & 0.032 & 0.692 & 0.662 & 0.499 & 30.318 & 0.975 \\
\textbf{RTRL+ChemDFM} & \textbf{0.535} & \textbf{\underline{45.550}} & \textbf{0.033} & \textbf{0.718} & \textbf{0.705} & \textbf{0.531} & \textbf{\underline{2.615}} & \textbf{0.986} \\ \midrule
  LlaSMol & 0.886 & 7.436 & 0.535 & 0.874 & 0.807 & 0.782 & 0.380 & 0.996 \\
  \textbf{RTRL+LlaSMol} & \textbf{0.912} & \textbf{\underline{5.298}} & \textbf{0.605} & \textbf{0.920} & \textbf{0.852} & \textbf{0.798} & \textbf{0.311} & \textbf{0.996} \\
\midrule
\rowcolor{gray!25} \multicolumn{9}{l}{\textit{USPTO-50 Reaction Prediction$\rightarrow$Retrosynthesis}} \\
  ChemDFM & 0.791 & 12.752 & 0.547 & 0.885 & 0.836 & 0.809 & \textbf{19.987} & 0.980 \\
\textbf{RTRL+ChemDFM} & \textbf{0.924} & \textbf{\underline{4.186}} & \textbf{\underline{0.821}} & \textbf{0.958} & \textbf{0.945} & \textbf{0.932} & 20.723 & \textbf{0.990} \\ \midrule
  LlaSMol & 0.810 & 17.117 & 0.304 & 0.827 & 0.743 & 0.702 & 0.644 & \textbf{0.997} \\
  \textbf{RTRL+LlaSMol} & \textbf{0.836} & \textbf{\underline{13.258}} & \textbf{0.352} & \textbf{0.851} & \textbf{0.762} & \textbf{0.726} & \textbf{0.512} & 0.995 \\
\midrule
\rowcolor{gray!25} \multicolumn{9}{l}{\textit{CHEBI-20 Molecule Captioning$\rightarrow$Text-based Molecule Generation}} \\
  ChemDFM & 0.598 & 43.747 & 0.170 & 0.723 & 0.591 & 0.504 & 28.474 & \textbf{0.983} \\
\textbf{RTRL+ChemDFM} & \textbf{0.693} & \textbf{\underline{31.831}} & \textbf{\underline{0.239}} & \textbf{0.795} & \textbf{0.668} & \textbf{0.598} & \textbf{\underline{1.724}} & 0.962 \\ \midrule
  LlaSMol & 0.586 & 40.163 & 0.041 & 0.670 & 0.468 & 0.386 & 8.706 & 0.956 \\
  \textbf{RTRL+LlaSMol} & \textbf{0.629} & \textbf{33.506} & \textbf{\underline{0.105}} & \textbf{0.673} & \textbf{0.487} &\textbf{0.411} & \textbf{\underline{5.438}} & \textbf{0.970} \\
\bottomrule
\end{tabular}}
\vspace{-10pt}
\caption{Round-trip consistency improvement by RTRL. \textbf{Bold} marks the best value per column; \underline{underline} marks an RTRL result that improves the corresponding base model by at least 20\% relative.}\label{tab:rtc}
\vspace{-10pt}
\end{table*}
\begin{table*}[t]
\centering
\resizebox{1.0\textwidth}{!}{
\begin{tabular}{lcccccccc}
\toprule
Task / Model & BLEU $\uparrow$ & Lev. $\downarrow$ & Exact Match $\uparrow$ & MACCS SIM. $\uparrow$ & RDKit SIM. $\uparrow$ & Morgan SIM. $\uparrow$ & FCD $\downarrow$ & Validity $\uparrow$ \\
\midrule
\rowcolor{gray!25} \multicolumn{9}{l}{\textit{CHEBI-20 Text-based Molecule Generation}} \\
  GPT (0-shot) & 0.475 & 48.985 & 0.164 & 0.605 & 0.475 & 0.441 & 20.837 & 0.687 \\
  Qwen-8B & 0.030 & 665.894 & 0.005 & 0.240 & 0.130 & 0.101 & 29.719 & 0.468 \\
  Mol-Instructions & 0.601 & 42.204 & 0.144 & 0.764 & 0.573 & 0.477 & 6.700 & \textbf{0.999} \\
  LlaSMol & 0.701 & 33.813 & 0.185 & 0.754 & 0.576 & 0.517 & 6.801 & 0.927 \\
  ChemDFM & 0.846 & 15.652 & 0.546 & 0.898 & 0.779 & 0.733 & \textbf{2.088} & 0.978 \\ \midrule
  \textbf{RTRL+LlaSMol} & 0.734 & 30.148 & 0.195 & 0.776 & 0.574 & 0.533 & 5.767 & 0.981 \\
\textbf{RTRL+ChemDFM} & \textbf{0.852} & \textbf{15.022} & \textbf{0.553} & \textbf{0.901} & \textbf{0.781} & \textbf{0.735} & 2.099 & 0.982 \\
\midrule
\rowcolor{gray!25} \multicolumn{9}{l}{\textit{USPTO-Mixed Reaction Prediction}} \\
  GPT (0-shot) & 0.691 & 16.194 & 0.433 & 0.819 & 0.768 & 0.727 & 5.968 & 0.910 \\
  Qwen-8B & 0.231 & 62.644 & 0.005 & 0.368 & 0.271 & 0.251 & 21.436 & 0.942 \\
  Mol-Instructions & 0.307 & 28.725 & 0.096 & 0.578 & 0.436 & 0.385 & 4.193 & \textbf{0.999} \\
  LlaSMol & 0.923 & 5.042 & 0.690 & 0.919 & 0.879 & 0.858 & 0.235 & 0.996 \\
  ChemDFM & 0.845 & 8.685 & 0.559 & 0.880 & 0.831 & 0.803 & 18.818 & 0.987 \\ \midrule
  \textbf{RTRL+LlaSMol} & \textbf{0.943} & \textbf{4.467} & \textbf{0.711} & \textbf{0.930} & \textbf{0.890} & \textbf{0.861} & 0.789 & 0.996 \\
\textbf{RTRL+ChemDFM} & 0.857 & 7.934 & 0.601 & 0.895 & 0.851 & 0.823 & \textbf{\underline{0.171}} & 0.985 \\
\midrule
\rowcolor{gray!25} \multicolumn{9}{l}{\textit{USPTO-50K Retrosynthesis Supervised}} \\
  GPT (10-shot) & 0.674 & 24.179 & 0.104 & 0.687 & 0.524 & 0.513 & 13.533 & 0.896 \\
  Qwen-8B & 0.486 & 46.429 & 0.015 & 0.739 & 0.663 & 0.546 & 23.224 & 0.944 \\
  Mol-Instructions & 0.588 & 36.660 & 0.004 & 0.697 & 0.540 & 0.474 & 12.352 & \textbf{0.999} \\
  LlaSMol & 0.818 & 16.027 & 0.349 & 0.845 & 0.773 & 0.734 & 0.579 & 0.999 \\
  ChemDFM & 0.802 & 17.249 & 0.311 & 0.840 & 0.766 & 0.728 & 0.293 & 0.990 \\
  GRPO+ChemDFM & 0.803 & 17.242 & 0.311 & 0.843 & 0.769 & 0.730 & \textbf{0.288} & 0.992 \\ \midrule
  \textbf{RTRL+LlaSMol} & \textbf{0.825} & \textbf{15.178} & \textbf{0.361} & 0.840 & 0.777 & 0.734 & 0.494 & 0.998 \\
\textbf{RTRL+ChemDFM} & 0.810 & 17.158 & 0.323 & \textbf{0.848} & \textbf{0.778} & \textbf{0.738} & 0.301 & 0.990 \\
\bottomrule
\end{tabular}}
\vspace{-10pt}
\caption{RTRL performance on molecule-based tasks. Highlighting as in Table~\ref{tab:rtc}.}
\label{tab:ss-mol}
\vspace{-10pt}
\end{table*}
We strive to answer the following research questions: \textbf{Q1:} Does RTRL actually improve round-trip consistency. \textbf{Q2:} Does enforcing round-trip consistency improve the model performance in both self-supervised and supervised cases? \textbf{Q3:} Does RTRL improve the generalizability of the base model?  \textbf{Q4:} Does iterative RTRL bring an extra performance boost, and when does the model stop improving? \textbf{Q5:} Can we iteratively improve the model from a seed dataset? \textbf{Q6:} How does RTRL compare to methods that can also do self-supervised learning, such as entropy minimization~\citep{agarwal2025unreasonable} and round-trip sample augmentation~\citep{tetko2020state}? We answer \textbf{Q1-5} in this section, and answer \textbf{Q6} in Appendix~\ref{app:moreexp}. We also include qualitative examples in Appendix~\ref{app:example}. Details for experiments, codes, and reproduction can be found in Appendix~\ref{app:exp}.

\textbf{Datasets and evaluations:} For molecule captioning and text-based molecule generation, we use CHEBI-20~\citep{edwards-etal-2021-text2mol} and Language-Plus-Molecule-24 (LM-24)~\citep{edwards-etal-2024-l}. For reaction prediction and retrosynthesis, we use USPTO-Mixed ~\citep{jin2017predicting} and USPTO-50K datasets~\citep{upsto50}, respectively. Detailed description of each dataset can be found in Appendix~\ref{app:data}. We follow the convention in existing chemical LLM work~\citep{edwards2022translation} to evaluate text and molecules. A detailed description of the evaluation metrics are in Appendix~\ref{app:exp}. All comparison are conducted after canonicalization of SMILES strings. In our result table, we use upward (downward) arrows to indicate that the metric is higher (lower) the better.

\textbf{Baselines:} We use Close-sourced LLM (GPT~\citep{openai2024gpt4technicalreport}), open-weight LLM (Qwen3-8B~\citep{yang2025qwen3technicalreport}), strong chemical LLMs (Mol-Instructions~\citep{fangmol}, LlaSMol~\citep{yullasmol} and ChemDFM~\citep{zhao2025developing}) as our baselines to cover a wide spectrum of chemically-capable LLMs. A special property of SMILES is its non-uniqueness, one molecule can corresponds to multiple valid SMILES. RTRL's behavior to this is largely dependent on the baseline LLM, LlaSMol is trained on canonicalized SMILES, while ChemDFM is trained on diverse non-canonicalized SMILES. We apply RTRL to both of them to show that \textbf{RTRL generalize across base models}.

\textbf{Round-trip consistency:} To answer \textbf{Q1}, we train base models with RTRL using only one domain of data (the data are unlabelled), and evaluate round-trip consistency change. After training, the consistency evaluation is performed on the test dataset. The results are in Table~\ref{tab:rtc}. We see improved round-trip consistency on all round-trip generation tasks. In particular, we see 43\% and 52\% relative exact match improvement in molecule captioning to generation and reaction prediction to retrosynthesis tasks for ChemDFM model. This validates our design to use an RL to improve the base model's RTC in bidirectional tasks \textbf{with only data from one domain} and \textbf{using a base model's own knowledge}.

\begin{table*}[t]
\centering

\resizebox{0.9\textwidth}{!}{
\begin{tabular}{lcccccc}
\toprule
Task / Model & BLEU-2 $\uparrow$ & BLEU-4 $\uparrow$ & ROUGE-1 $\uparrow$ & ROUGE-2 $\uparrow$ & ROUGE-L $\uparrow$ & METEOR $\uparrow$ \\
\midrule
\rowcolor{gray!25} \multicolumn{7}{l}{\textit{CHEBI-20}} \\
  GPT (0-shot) & 0.026 & 0.007 & 0.086 & 0.023 & 0.055 & 0.173 \\
  Qwen-8B & 0.019 & 0.004 & 0.099 & 0.025 & 0.070 & 0.138 \\
  Mol-Instructions & 0.076 & 0.056 & 0.249 & 0.175 & 0.236 & 0.164 \\
  LlaSMol & 0.386 & 0.275 & 0.490 & 0.306 & 0.423 & 0.419 \\
  ChemDFM & 0.286 & 0.244 & 0.406 & 0.312 & 0.378 & 0.345 \\ \midrule
  \textbf{RTRL+LlaSMol} & 0.434 & 0.290 & \textbf{0.535} & 0.324 & 0.449 & 0.440 \\
\textbf{RTRL+ChemDFM} & \textbf{\underline{0.447}} & \textbf{\underline{0.380}} & \underline{0.529} & \textbf{\underline{0.406}} & \textbf{\underline{0.483}} & \textbf{\underline{0.481}} \\
\midrule
\rowcolor{gray!25} \multicolumn{7}{l}{\textit{CHEBI-20 Supervised}} \\
  GPT (10-shot) & 0.042 & 0.015 & 0.131 & 0.040 & 0.089 & 0.231 \\
  Qwen-8B & 0.163 & 0.098 & 0.398 & 0.219 & 0.341 & 0.325 \\
  Mol-Instructions & 0.434 & 0.329 & 0.529 & 0.358 & 0.467 & 0.466 \\
  ChemDFM & 0.385 & 0.333 & 0.493 & 0.388 & 0.457 & 0.437 \\
  GRPO+ChemDFM & 0.418 & 0.358 & 0.523 & 0.397 & 0.475 & 0.457 \\ \midrule
\textbf{RTRL+ChemDFM} & \textbf{\underline{0.492}} & \textbf{\underline{0.421}} & \textbf{0.563} & \textbf{0.437} & \textbf{0.515} & \textbf{0.521} \\
\midrule
\rowcolor{gray!25} \multicolumn{7}{l}{\textit{LM-24 Supervised}} \\
  GPT (10-shot) & 0.013 & 0.004 & 0.056 & 0.018 & 0.040 & 0.114 \\
  Qwen-8B & 0.745 & 0.537 & 0.763 & 0.572 & 0.550 & 0.706 \\
  Mol-Instructions & 0.764 & \textbf{0.554} & 0.782 & 0.589 & 0.562 & 0.727 \\
  ChemDFM & 0.752 & 0.545 & 0.779 & 0.586 & 0.562 & 0.720 \\
  GRPO+ChemDFM & 0.734 & 0.534 & 0.748 & 0.555 & 0.548 & 0.708 \\ \midrule
\textbf{RTRL+ChemDFM} & \textbf{0.766} & 0.551 & \textbf{0.790} & \textbf{0.599} & \textbf{0.567} & \textbf{0.735} \\
\bottomrule
\end{tabular}}
\vspace{-5pt}
\caption{RTRL performance on text-based tasks. Highlighting as in Table~\ref{tab:rtc}.}
\label{tab:ss-text}
\vspace{-15pt}
\end{table*}


\textbf{Round-trip consistency and model performance:} Simply maximizing consistency is not sufficient. Hence, we need to answer \textbf{Q2} by studying the performance of the RTRL-trained models. In the self-supervised scenario, we only have access to one domain of data. The molecule-based results are in Table~\ref{tab:ss-mol}, and the text-based results are in Table~\ref{tab:ss-text}. We can see the RTRL-trained model outperform the base model on almost every task. RTRL achieved 7.5\% relative improvement in exact match on the reaction prediction task (0.601 vs 0.559) using ChemDFM as base model. It also achieves at least 27\% relative improvement in all metrics on the CHEBI-20 molecule captioning task. Similarly, using LlaSMol as base model, RTRL improve it on 7 out of 8 targets. These findings confirm our central goal to use RTC to improve the model's coherence and performance. Conventionally, model requires access to data from both domain, while RTRL \textbf{liberates} the model by requiring only one domain of data, which is prevalent and much easier to acquire in the chemical domain.

Gains are largest in the self-supervised and captioning settings and more modest on supervised USPTO-50K retrosynthesis, where high-quality verifiable rewards already exist and the round-trip reward mainly adds a regularization effect. In the supervised setting, we perform an SFT on the bidirectional tasks before RL. During RL training, we add the evaluation metric score as an auxiliary score to fully utilize the label. We use (a subset of) the finetuning data in RL to ensure that the benefit does come from extra data. For the baselines, we compared the same set of LLMs as in the self-supervised scenario, but they are also finetuned with the same data. We additionally compared to GRPO+ChemDFM, where only the metric scores are used as a reward in the GRPO. The molecule-based results are in Table~\ref{tab:ss-mol}, and the text-based results are in Table~\ref{tab:ss-text}. In this scenario, we still see improvement on most tasks, showing effective self-improvement. Additional supervised and self-supervised results can be found in Appendix~\ref{app:moreexp}.

\textbf{Is the improvement from the round-trip reward, or just GRPO?} We address this in two parts. First, under the \emph{same} training setup and reward budget, the round-trip reward yields consistent gains over the verifiable-metric-only GRPO+ChemDFM baseline (e.g., 0.492 vs. 0.418 BLEU-2 on CHEBI-20 supervised captioning in Table~\ref{tab:ss-text}; 0.547 vs. 0.504 exact match on supervised molecule generation in Table~\ref{tab:more_mol}), showing it provides \emph{complementary} signal. Second, GRPO with verifiable rewards is \textbf{inapplicable} in the self-supervised setting, where metric rewards (MACCS, RDKit, Morgan) cannot be computed without ground-truth references. All self-supervised results (Tables~\ref{tab:rtc},~\ref{tab:ss-mol}, and the CHEBI-20 section of Table~\ref{tab:ss-text}) are thus gains GRPO alone cannot achieve---the label-free round-trip reward is precisely RTRL's unique contribution.

\begin{table*}[t]
\centering
\resizebox{\linewidth}{!}{
\begin{tabular}{lccccccccc}
\toprule
Model & Train Time & BLEU $\uparrow$ & Lev. $\downarrow$ & Exact $\uparrow$ & MACCS $\uparrow$ & RDKit $\uparrow$ & Morgan $\uparrow$ & FCD $\downarrow$ & Validity $\uparrow$ \\
\midrule
RTMol & 9h 14m & 0.827 & 17.491 & 0.538 & 0.893 & 0.757 & 0.714 & \textbf{1.936} & 0.980 \\
\textbf{RTRL} & \textbf{5h 37m} & \textbf{0.852} & \textbf{15.022} & \textbf{0.553} & \textbf{0.901} & \textbf{0.781} & \textbf{0.735} & 2.099 & \textbf{0.982} \\
\bottomrule
\end{tabular}}

\vspace{-8pt}
\caption{Head-to-head comparison to the concurrent RTMol~\citep{rtmol} on CHEBI-20 text-based molecule generation. Both methods use the \emph{same} ChemDFM backbone, training data, and round-trip reward budget; training time is measured on identical hardware.}\label{tab:rtmol}
\vspace{-12pt}
\end{table*}
\textbf{Comparison to RTMol:} The concurrent RTMol~\citep{rtmol} also uses round-trip rewards for chemical LLMs. Under a matched setup---same ChemDFM backbone, training data, and reward budget---on CHEBI-20 text-based molecule generation, RTRL attains comparable or better accuracy on all metrics except FCD while requiring \textbf{39\% less training time} (Table~\ref{tab:rtmol}), as its single-forward-pass log-likelihood reward avoids RTMol's full autoregressive backward generation. RTRL additionally offers iterative mutual improvement and a task-agnostic reward, whereas RTMol's Tanimoto reward is specific to molecule outputs.

\begin{wraptable}{r}{6cm}

\resizebox{\linewidth}{!}{
\begin{tabular}{lcccccc}
\toprule
Model & ESOL $\downarrow$ & Lipo $\downarrow$ & BBBP $\uparrow$ & Clintox $\uparrow$ \\
\midrule
Llasmol & 1.438 & 1.09 & 74.1 & 91.8  \\
\textbf{Llasmol+RTRL} & \textbf{1.359} & \textbf{1.01} & \textbf{74.6} & \textbf{93.1}  \\
\bottomrule
\end{tabular}}
\vspace{-10pt}
\caption{RTRL transferability.}\label{tab:mol-trans}
\vspace{-10pt}
\end{wraptable}
\textbf{Q3} examines the generalizability of RTRL. In particular, when RTRL enhances a pair of bidirectional tasks, does it improve the base model's inherent understanding of the subject? 
We present the results when LlaSMol is RTRL-trained with Retrosynthesis-reaction-prediction task and then applied to a set of property prediction tasks in Table~\ref{tab:mol-trans}. While these tasks are not explicitly optimized by RTRL, RTRL improves the model's coherence in chemical knowledge, and the model's performance on these tasks also improves. Appendix~\ref{app:moreexp} includes results where RTRL is directly applied to classification tasks and shows effectiveness.

\begin{figure*}[t]
    \centering 

    \begin{minipage}{0.247\textwidth}
        \includegraphics[width=\linewidth, trim={0.6cm 0.6cm 0.6cm 0.6cm}, clip]{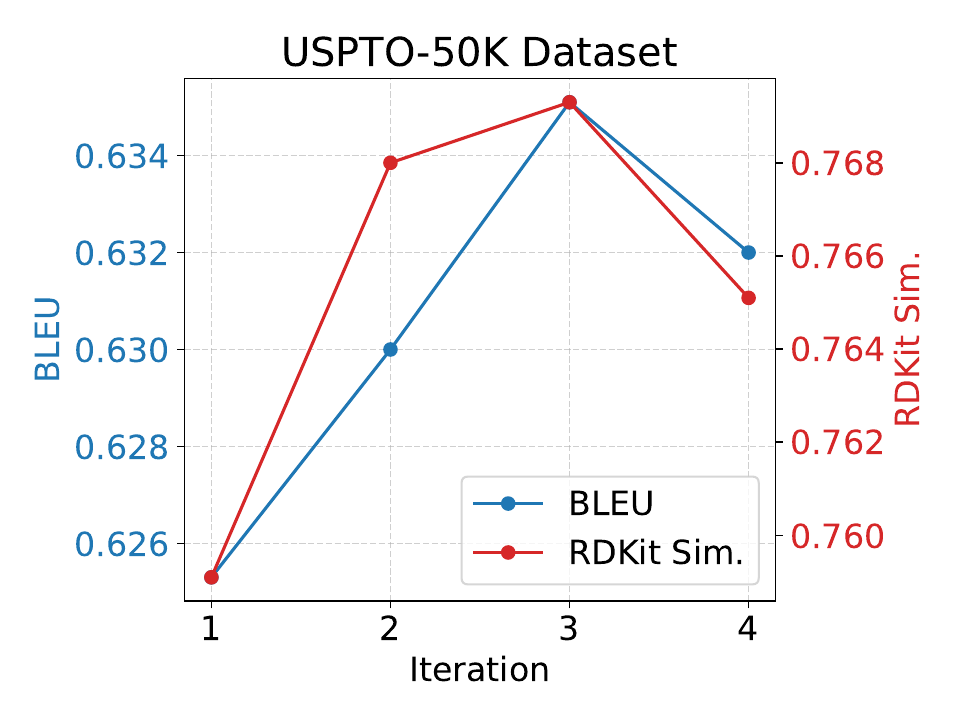}
    \end{minipage}\hfill %
    \begin{minipage}{0.247\textwidth}
        \includegraphics[width=\linewidth, trim={0.6cm 0.6cm 0.6cm 0.6cm}, clip]{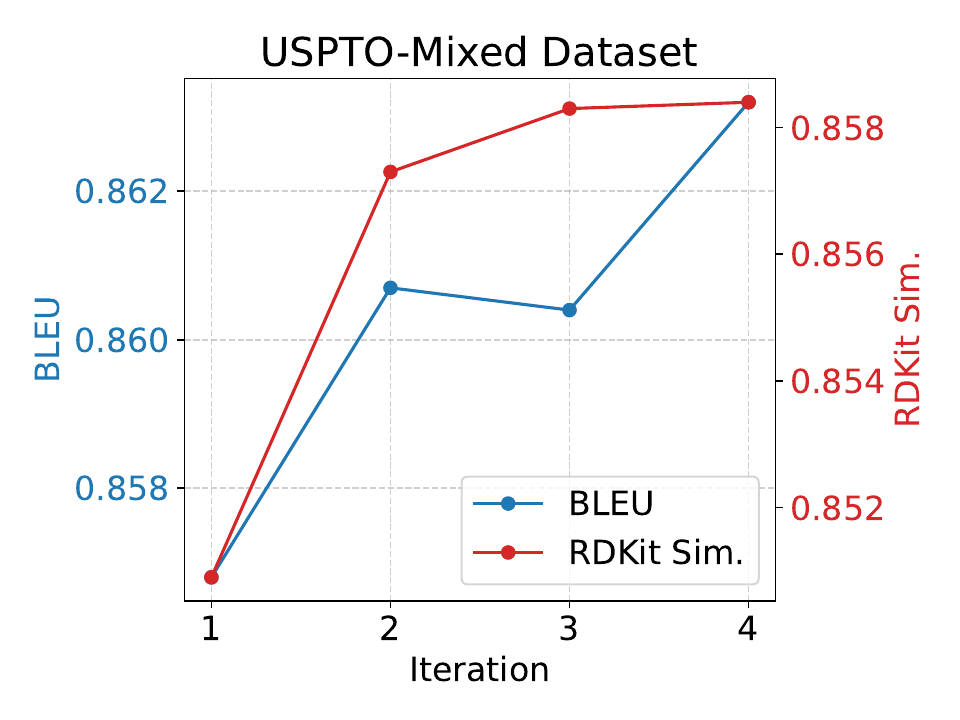}
    \end{minipage}\hfill %
    \begin{minipage}{0.247\textwidth}
        \includegraphics[width=\linewidth, trim={0.6cm 0.6cm 0.6cm 0.6cm}, clip]{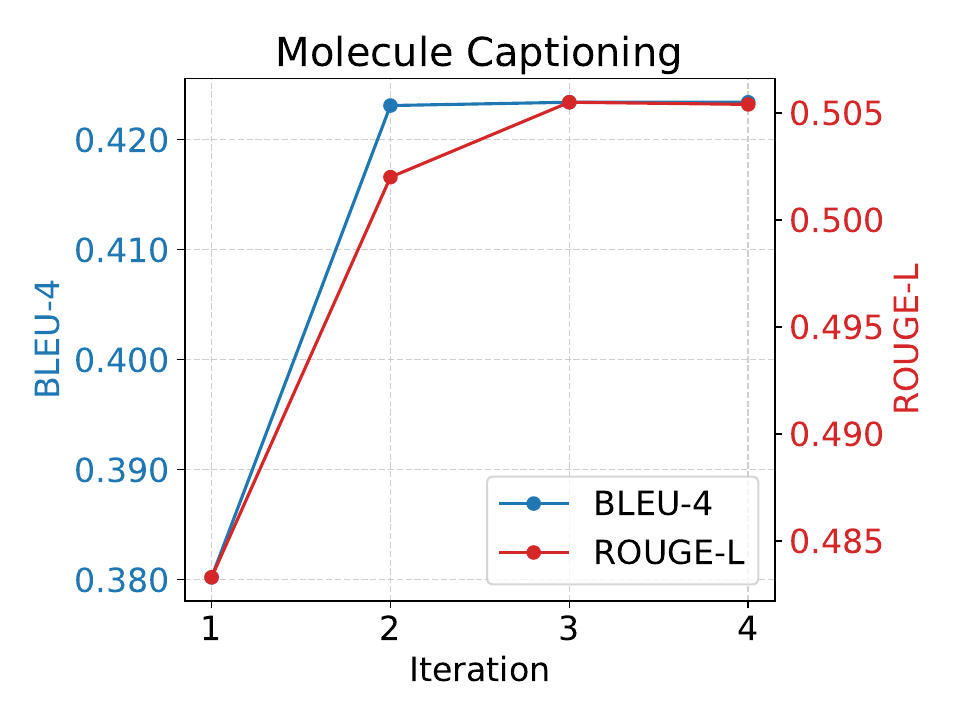}
    \end{minipage}\hfill %
    \begin{minipage}{0.247\textwidth}
        \includegraphics[width=\linewidth, trim={0.6cm 0.6cm 0.6cm 0.6cm}, clip]{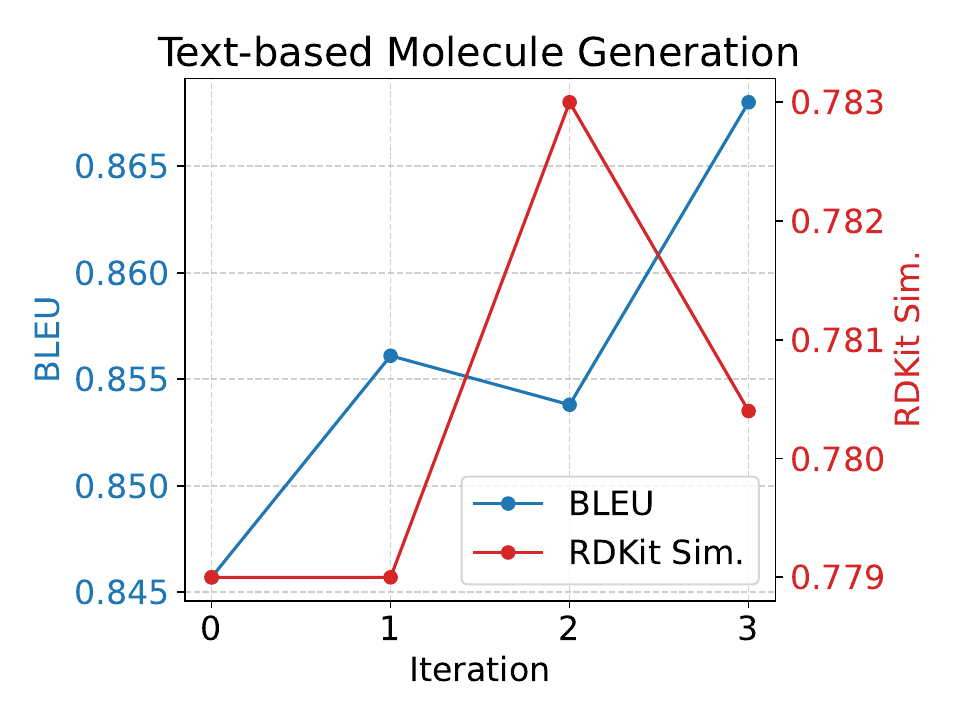}
    \end{minipage}

    \caption{Model performance progression in iterative RTRL. From left to right: (1) Retrosynthesis performance. Supervised. (2) Reaction prediction. Supervised. (3) CHEBI-20. Synthetic. (4) CHEBI-20. Synthetic. Iteration 0 means the base model.}
    \label{fig:imp}
\vspace{-10pt}
\end{figure*}

\textbf{Iterative RTRL:} To answer \textbf{Q4}, we apply RTRL to USPTO-50 and USPTO-Mixed dataset. They are a pair of bidirectional tasks, but their data entries do not match. We repeat the RTRL process 4 times and record the model performance. We draw a line plot of BLEU and RDKit Similarity in Figure~\ref{fig:imp}. We see that iterative RL can continuously improve the model's ability in both directions. This greatly extends the applicable scenario and effectiveness of RTRL to accommodate different data availability.

\textbf{Synthetic RTRL:} To answer \textbf{Q5}, we apply RTRL to CHEBI-20 with caption removed as the seed dataset. We swap the roles of forward and backward functions and generate synthetic data from the seed dataset. Then, we use RTRL on this synthetic data to train the forward (previously backward) function. We repeat and plot performance on both directions in Figure~\ref{fig:imp}. We can see clear improvement for both directions even without a ground-truth label from one domain. Meanwhile, the improvement is less consistent, potentially because of the variability data sampling. We leave designing a more controlled synthesis process to future work.

\section{Conclusion, Limitation, and Future Work}
In this paper, we highlight the critical role of round-trip consistency (RTC) in LLMs and propose Round-Trip Reinforcement Learning (RTRL) to algorithmically enforce it via GRPO. Empirical results across diverse chemical tasks demonstrate significant, consistent improvements over strong baselines, validating RTRL's efficacy.

We validate RTRL within computational chemistry. Its core prerequisites---bidirectional tasks with mappings that can be inverted, and a base model with initial competence in both directions---are not specific to chemistry, and we outline candidate domains such as code migration and low-resource translation in Appendix~\ref{app:rtrl_usecase}; validating RTRL in those settings is left to future work. Fully realizing RTRL's potential also involves addressing several broader challenges in generative modeling. To rigorously isolate the impact of RTC as a trainable objective, our current framework leaves three primary avenues for future work: (1) it relies on external seed data rather than operating in completely synthetic scenarios; (2) its likelihood objective prioritizes highly reliable pathways over diversity in one-to-many tasks; and (3) it uses a frozen backward model to maintain a strictly stationary reward landscape, leaving dynamic judge co-evolution unexplored. We provide a detailed discussion of these limitation and potential pathways to solve them in Appendix~\ref{app:limitation}.

Moving forward, we plan to address these challenges and establish RTRL as a foundational principle for bidirectional self-improvement.

\bibliography{colm2026_conference}

@article{zhao2025developing,
   title={Developing ChemDFM as a large language foundation model for chemistry},
   volume={6},
   ISSN={2666-3864},
   url={http://dx.doi.org/10.1016/j.xcrp.2025.102523},
   DOI={10.1016/j.xcrp.2025.102523},
   number={4},
   journal={Cell Reports Physical Science},
   publisher={Elsevier BV},
   author={Zhao, Zihan and Ma, Da and Chen, Lu and Sun, Liangtai and Li, Zihao and Xia, Yi and Chen, Bo and Xu, Hongshen and Zhu, Zichen and Zhu, Su and Fan, Shuai and Shen, Guodong and Yu, Kai and Chen, Xin},
   year={2025},
   month=apr, pages={102523} }

@misc{fangmol,
      title={Mol-Instructions: A Large-Scale Biomolecular Instruction Dataset for Large Language Models}, 
      author={Yin Fang and Xiaozhuan Liang and Ningyu Zhang and Kangwei Liu and Rui Huang and Zhuo Chen and Xiaohui Fan and Huajun Chen},
      year={2024},
      eprint={2306.08018},
      archivePrefix={arXiv},
      primaryClass={q-bio.QM},
      url={https://arxiv.org/abs/2306.08018}, 
}

@inproceedings{pei2024biot5+,
  title={BioT5+: Towards Generalized Biological Understanding with IUPAC Integration and Multi-task Tuning},
  author={Pei, Qizhi and Wu, Lijun and Gao, Kaiyuan and Liang, Xiaozhuan and Fang, Yin and Zhu, Jinhua and Xie, Shufang and Qin, Tao and Yan, Rui},
  booktitle={ACL (Findings)},
  year={2024}
}

@inproceedings{cao2025instructmol,
  title={InstructMol: Multi-Modal Integration for Building a Versatile and Reliable Molecular Assistant in Drug Discovery},
  author={Cao, He and Liu, Zijing and Lu, Xingyu and Yao, Yuan and Li, Yu},
  booktitle={Proceedings of the 31st International Conference on Computational Linguistics},
  pages={354--379},
  year={2025}
}

@article{wang2025survey,
  title={A Survey of Large Language Models for Text-Guided Molecular Discovery: from Molecule Generation to Optimization},
  author={Wang, Ziqing and Zhang, Kexin and Zhao, Zihan and Wen, Yibo and Pandey, Abhishek and Liu, Han and Ding, Kaize},
  journal={arXiv preprint arXiv:2505.16094},
  year={2025}
}

@article{liu2024evaluating,
  title={Evaluating Molecule Synthesizability via Retrosynthetic Planning and Reaction Prediction},
  author={Liu, Songtao and Zhang, Dandan and Tu, Zhengkai and Dai, Hanjun and Liu, Peng},
  journal={arXiv preprint arXiv:2411.08306},
  year={2024}
}

@inproceedings{edwards-etal-2024-l,
    title = "{L}+{M}-24: Building a Dataset for {L}anguage+{M}olecules @ {ACL} 2024",
    author = "Edwards, Carl  and
      Wang, Qingyun  and
      Zhao, Lawrence  and
      Ji, Heng",
    editor = "Edwards, Carl  and
      Wang, Qingyun  and
      Li, Manling  and
      Zhao, Lawrence  and
      Hope, Tom  and
      Ji, Heng",
    booktitle = "Proceedings of the 1st Workshop on Language + Molecules (L+M 2024)",
    month = aug,
    year = "2024",
    address = "Bangkok, Thailand",
    publisher = "Association for Computational Linguistics",
    url = "https://aclanthology.org/2024.langmol-1.1/",
    doi = "10.18653/v1/2024.langmol-1.1",
    pages = "1--9"
}

@inproceedings{pei2024enhanced,
  title={Enhanced BioT5+ for Molecule-Text Translation: A Three-Stage Approach with Data Distillation, Diverse Training, and Voting Ensemble},
  author={Pei, Qizhi and Wu, Lijun and Gao, Kaiyuan and Zhu, Jinhua and Yan, Rui},
  booktitle={Proceedings of the 1st Workshop on Language+ Molecules (L+ M 2024)},
  pages={48--54},
  year={2024}
}

@inproceedings{allamanis2024unsupervised,
  title={Unsupervised Evaluation of Code LLMs with Round-Trip Correctness},
  author={Allamanis, Miltiadis and Panthaplackel, Sheena and Yin, Pengcheng},
  booktitle={International Conference on Machine Learning},
  pages={1050--1066},
  year={2024},
  organization={PMLR}
}

@inproceedings{alberti2019synthetic,
  title={Synthetic QA Corpora Generation with Roundtrip Consistency},
  author={Alberti, Chris and Andor, Daniel and Pitler, Emily and Devlin, Jacob and Collins, Michael},
  booktitle={Proceedings of the 57th Annual Meeting of the Association for Computational Linguistics},
  pages={6168--6173},
  year={2019}
}

@inproceedings{rennie2020unsupervised,
  title={Unsupervised adaptation of question answering systems via generative self-training},
  author={Rennie, Steven and Marcheret, Etienne and Mallinar, Neil and Nahamoo, David and Goel, Vaibhava},
  booktitle={Proceedings of the 2020 Conference on Empirical Methods in Natural Language Processing (EMNLP)},
  pages={1148--1157},
  year={2020}
}

@article{hong2025consistencychecker,
  title={ConsistencyChecker: Tree-based Evaluation of LLM Generalization Capabilities},
  author={Hong, Zhaochen and Yu, Haofei and You, Jiaxuan},
  journal={arXiv preprint arXiv:2506.12376},
  year={2025}
}

@inproceedings{somers-2005-round,
    title = "Round-trip Translation: What Is It Good For?",
    author = "Somers, Harold",
    editor = "Baldwin, Timothy  and
      Curran, James  and
      van Zaanen, Menno",
    booktitle = "Proceedings of the Australasian Language Technology Workshop 2005",
    month = dec,
    year = "2005",
    address = "Sydney, Australia",
    url = "https://aclanthology.org/U05-1019/",
    pages = "127--133"
}

@inproceedings{yung2025round,
  title={Round trip translation defence against large language model jailbreaking attacks},
  author={Yung, Canaan and Dolatabadi, Hadi Mohaghegh and Erfani, Sarah and Leckie, Christopher},
  booktitle={Pacific-Asia Conference on Knowledge Discovery and Data Mining},
  pages={286--297},
  year={2025},
  organization={Springer}
}

@article{cheng2021molecular,
  title={Molecular design in drug discovery: a comprehensive review of deep generative models},
  author={Cheng, Yu and Gong, Yongshun and Liu, Yuansheng and Song, Bosheng and Zou, Quan},
  journal={Briefings in bioinformatics},
  volume={22},
  number={6},
  pages={bbab344},
  year={2021},
  publisher={Oxford University Press}
}

@article{yang2024molecule,
  title={Molecule generation for drug design: a graph learning perspective},
  author={Yang, Nianzu and Wu, Huaijin and Zeng, Kaipeng and Li, Yang and Bao, Siyuan and Yan, Junchi},
  journal={Fundamental Research},
  year={2024},
  publisher={Elsevier}
}

@misc{yullasmol,
      title={LlaSMol: Advancing Large Language Models for Chemistry with a Large-Scale, Comprehensive, High-Quality Instruction Tuning Dataset}, 
      author={Botao Yu and Frazier N. Baker and Ziqi Chen and Xia Ning and Huan Sun},
      year={2024},
      eprint={2402.09391},
      archivePrefix={arXiv},
      primaryClass={cs.AI},
      url={https://arxiv.org/abs/2402.09391}, 
}

@article{zhang2024chemllm,
  title={Chemllm: A chemical large language model},
  author={Zhang, Di and Liu, Wei and Tan, Qian and Chen, Jingdan and Yan, Hang and Yan, Yuliang and Li, Jiatong and Huang, Weiran and Yue, Xiangyu and Ouyang, Wanli and others},
  journal={arXiv preprint arXiv:2402.06852},
  year={2024}
}

@inproceedings{dey-etal-2025-mathtt,
    title = "{GeLLM{\textthreesuperior}O}: Generalizing Large Language Models for Multi-property Molecule Optimization",
    author = "Dey, Vishal  and
      Hu, Xiao  and
      Ning, Xia",
    editor = "Che, Wanxiang  and
      Nabende, Joyce  and
      Shutova, Ekaterina  and
      Pilehvar, Mohammad Taher",
    booktitle = "Proceedings of the 63rd Annual Meeting of the Association for Computational Linguistics (Volume 1: Long Papers)",
    month = jul,
    year = "2025",
    address = "Vienna, Austria",
    publisher = "Association for Computational Linguistics",
    url = "https://aclanthology.org/2025.acl-long.1225/",
    doi = "10.18653/v1/2025.acl-long.1225",
    pages = "25192--25221",
    ISBN = "979-8-89176-251-0"
}

@article{ye2025drugassist,
  title={Drugassist: A large language model for molecule optimization},
  author={Ye, Geyan and Cai, Xibao and Lai, Houtim and Wang, Xing and Huang, Junhong and Wang, Longyue and Liu, Wei and Zeng, Xiangxiang},
  journal={Briefings in Bioinformatics},
  volume={26},
  number={1},
  pages={bbae693},
  year={2025},
  publisher={Oxford University Press}
}

@inproceedings{edwards2022translation,
  title={Translation between Molecules and Natural Language},
  author={Edwards, Carl and Lai, Tuan and Ros, Kevin and Honke, Garrett and Cho, Kyunghyun and Ji, Heng},
  booktitle={Proceedings of the 2022 Conference on Empirical Methods in Natural Language Processing},
  pages={375--413},
  year={2022}
}

@article{li2024molreflect,
  title={MolReFlect: Towards In-Context Fine-grained Alignments between Molecules and Texts},
  author={Li, Jiatong and Liu, Yunqing and Liu, Wei and Le, Jingdi and Zhang, Di and Fan, Wenqi and Zhou, Dongzhan and Li, Yuqiang and Li, Qing},
  journal={arXiv preprint arXiv:2411.14721},
  year={2024}
}

@article{lin2024property,
  title={Property Enhanced Instruction Tuning for Multi-task Molecule Generation with Large Language Models},
  author={Lin, Xuan and Chen, Long and Wang, Yile and Zeng, Xiangxiang and Yu, Philip S},
  journal={arXiv preprint arXiv:2412.18084},
  year={2024}
}

@misc{guo2025unimotunifiedmoleculetextlanguage,
      title={UniMoT: Unified Molecule-Text Language Model with Discrete Token Representation}, 
      author={Shuhan Guo and Yatao Bian and Ruibing Wang and Nan Yin and Zhen Wang and Quanming Yao},
      year={2025},
      eprint={2408.00863},
      archivePrefix={arXiv},
      primaryClass={cs.CL},
      url={https://arxiv.org/abs/2408.00863}, 
}

@misc{xia2025naturelanguagemodeldeciphering,
      title={Nature Language Model: Deciphering the Language of Nature for Scientific Discovery}, 
      author={Yingce Xia and Peiran Jin and Shufang Xie and Liang He and Chuan Cao and Renqian Luo and Guoqing Liu and Yue Wang and Zequn Liu and Yuan-Jyue Chen and Zekun Guo and Yeqi Bai and Pan Deng and Yaosen Min and Ziheng Lu and Hongxia Hao and Han Yang and Jielan Li and Chang Liu and Jia Zhang and Jianwei Zhu and Ran Bi and Kehan Wu and Wei Zhang and Kaiyuan Gao and Qizhi Pei and Qian Wang and Xixian Liu and Yanting Li and Houtian Zhu and Yeqing Lu and Mingqian Ma and Zun Wang and Tian Xie and Krzysztof Maziarz and Marwin Segler and Zhao Yang and Zilong Chen and Yu Shi and Shuxin Zheng and Lijun Wu and Chen Hu and Peggy Dai and Tie-Yan Liu and Haiguang Liu and Tao Qin},
      year={2025},
      eprint={2502.07527},
      archivePrefix={arXiv},
      primaryClass={cs.AI},
      url={https://arxiv.org/abs/2502.07527}, 
}

@article{jang2024can,
  title={Can LLMs Generate Diverse Molecules? Towards Alignment with Structural Diversity},
  author={Jang, Hyosoon and Jang, Yunhui and Kim, Jaehyung and Ahn, Sungsoo},
  journal={arXiv preprint arXiv:2410.03138},
  year={2024}
}

@article{calanzone2025mol,
  title={Mol-MoE: Training Preference-Guided Routers for Molecule Generation},
  author={Calanzone, Diego and D'Oro, Pierluca and Bacon, Pierre-Luc},
  journal={arXiv preprint arXiv:2502.05633},
  year={2025}
}

@article{gkoumas2024almol,
  title={ALMol: Aligned Language-Molecule Translation LLMs through Offline Preference Contrastive Optimisation},
  author={Gkoumas, Dimitris},
  journal={arXiv preprint arXiv:2405.08619},
  year={2024}
}

@article{lee2025mol,
  title={Mol-LLM: Multimodal Generalist Molecular LLM with Improved Graph Utilization},
  author={Lee, Chanhui and Ko, Hanbum and Song, Yuheon and Jeong, YongJun and Hormazabal, Rodrigo and Han, Sehui and Bae, Kyunghoon and Lim, Sungbin and Kim, Sungwoong},
  journal={arXiv preprint arXiv:2502.02810},
  year={2025}
}

@misc{guanrstar,
      title={rStar-Math: Small LLMs Can Master Math Reasoning with Self-Evolved Deep Thinking}, 
      author={Xinyu Guan and Li Lyna Zhang and Yifei Liu and Ning Shang and Youran Sun and Yi Zhu and Fan Yang and Mao Yang},
      year={2025},
      eprint={2501.04519},
      archivePrefix={arXiv},
      primaryClass={cs.CL},
      url={https://arxiv.org/abs/2501.04519}, 
}

@article{huang2025r,
  title={R-Zero: Self-Evolving Reasoning LLM from Zero Data},
  author={Huang, Chengsong and Yu, Wenhao and Wang, Xiaoyang and Zhang, Hongming and Li, Zongxia and Li, Ruosen and Huang, Jiaxin and Mi, Haitao and Yu, Dong},
  journal={arXiv preprint arXiv:2508.05004},
  year={2025}
}

@article{ma2025s,
  title={$S^2R$: Teaching LLMs to Self-verify and Self-correct via Reinforcement Learning},
  author={Ma, Ruotian and Wang, Peisong and Liu, Cheng and Liu, Xingyan and Chen, Jiaqi and Zhang, Bang and Zhou, Xin and Du, Nan and Li, Jia},
  journal={arXiv preprint arXiv:2502.12853},
  year={2025}
}

@article{gao2025survey,
  title={A survey of self-evolving agents: On path to artificial super intelligence},
  author={Gao, Huan-ang and Geng, Jiayi and Hua, Wenyue and Hu, Mengkang and Juan, Xinzhe and Liu, Hongzhang and Liu, Shilong and Qiu, Jiahao and Qi, Xuan and Wu, Yiran and others},
  journal={arXiv preprint arXiv:2507.21046},
  year={2025}
}

@article{shao2025spurious,
  title={Spurious rewards: Rethinking training signals in rlvr},
  author={Shao, Rulin and Li, Shuyue Stella and Xin, Rui and Geng, Scott and Wang, Yiping and Oh, Sewoong and Du, Simon Shaolei and Lambert, Nathan and Min, Sewon and Krishna, Ranjay and others},
  journal={arXiv preprint arXiv:2506.10947},
  year={2025}
}

@inproceedings{huang2023large,
  title={Large Language Models Can Self-Improve},
  author={Huang, Jiaxin and Gu, Shixiang and Hou, Le and Wu, Yuexin and Wang, Xuezhi and Yu, Hongkun and Han, Jiawei},
  booktitle={Proceedings of the 2023 Conference on Empirical Methods in Natural Language Processing},
  pages={1051--1068},
  year={2023}
}

@article{agarwal2025unreasonable,
  title={The unreasonable effectiveness of entropy minimization in llm reasoning},
  author={Agarwal, Shivam and Zhang, Zimin and Yuan, Lifan and Han, Jiawei and Peng, Hao},
  journal={arXiv preprint arXiv:2505.15134},
  year={2025}
}

@article{shao2024deepseekmath,
  title={Deepseekmath: Pushing the limits of mathematical reasoning in open language models},
  author={Shao, Zhihong and Wang, Peiyi and Zhu, Qihao and Xu, Runxin and Song, Junxiao and Bi, Xiao and Zhang, Haowei and Zhang, Mingchuan and Li, YK and others},
  journal={arXiv preprint arXiv:2402.03300},
  year={2024}
}

@inproceedings{edwards-etal-2021-text2mol,
    title = "{T}ext2{M}ol: Cross-Modal Molecule Retrieval with Natural Language Queries",
    author = "Edwards, Carl  and
      Zhai, ChengXiang  and
      Ji, Heng",
    editor = "Moens, Marie-Francine  and
      Huang, Xuanjing  and
      Specia, Lucia  and
      Yih, Scott Wen-tau",
    booktitle = "Proceedings of the 2021 Conference on Empirical Methods in Natural Language Processing",
    month = nov,
    year = "2021",
    address = "Online and Punta Cana, Dominican Republic",
    publisher = "Association for Computational Linguistics",
    url = "https://aclanthology.org/2021.emnlp-main.47/",
    doi = "10.18653/v1/2021.emnlp-main.47",
    pages = "595--607"
}

@article{jin2017predicting,
  title={Predicting Organic Reaction Outcomes with Weisfeiler-Lehman Network},
  author={Jin, Wengong and Coley, Connor and Barzilay, Regina and Jaakkola, Tommi},
  journal={Advances in Neural Information Processing Systems},
  volume={30},
  year={2017}
}

@article{upsto50,
author = {Schneider, Nadine and Stiefl, Nikolaus and Landrum, Gregory A.},
title = {What’s What: The (Nearly) Definitive Guide to Reaction Role Assignment},
journal = {Journal of Chemical Information and Modeling},
volume = {56},
number = {12},
pages = {2336-2346},
year = {2016},
doi = {10.1021/acs.jcim.6b00564},
note ={PMID: 28024398},
URL = {
        https://doi.org/10.1021/acs.jcim.6b00564
},
eprint = {
        https://doi.org/10.1021/acs.jcim.6b00564
}
}

@misc{yang2025qwen3technicalreport,
      title={Qwen3 Technical Report}, 
      author={An Yang and Anfeng Li and Baosong Yang and Beichen Zhang and Binyuan Hui and Bo Zheng and Bowen Yu and Chang Gao and Chengen Huang and Chenxu Lv and Chujie Zheng and Dayiheng Liu and Fan Zhou and Fei Huang and Feng Hu and Hao Ge and Haoran Wei and Huan Lin and Jialong Tang and Jian Yang and Jianhong Tu and Jianwei Zhang and Jianxin Yang and Jiaxi Yang and Jing Zhou and Jingren Zhou and Junyang Lin and Kai Dang and Keqin Bao and Kexin Yang and Le Yu and Lianghao Deng and Mei Li and Mingfeng Xue and Mingze Li and Pei Zhang and Peng Wang and Qin Zhu and Rui Men and Ruize Gao and Shixuan Liu and Shuang Luo and Tianhao Li and Tianyi Tang and Wenbiao Yin and Xingzhang Ren and Xinyu Wang and Xinyu Zhang and Xuancheng Ren and Yang Fan and Yang Su and Yichang Zhang and Yinger Zhang and Yu Wan and Yuqiong Liu and Zekun Wang and Zeyu Cui and Zhenru Zhang and Zhipeng Zhou and Zihan Qiu},
      year={2025},
      eprint={2505.09388},
      archivePrefix={arXiv},
      primaryClass={cs.CL},
      url={https://arxiv.org/abs/2505.09388}, 
}

@misc{openai2024gpt4technicalreport,
      title={GPT-4 Technical Report}, 
      author={OpenAI and Josh Achiam and Steven Adler and Sandhini Agarwal and Lama Ahmad and Ilge Akkaya and Florencia Leoni Aleman and Diogo Almeida and Janko Altenschmidt and Sam Altman and Shyamal Anadkat and Red Avila and Igor Babuschkin and Suchir Balaji and Valerie Balcom and Paul Baltescu and Haiming Bao and Mohammad Bavarian and Jeff Belgum and Irwan Bello and Jake Berdine and Gabriel Bernadett-Shapiro and Christopher Berner and Lenny Bogdonoff and Oleg Boiko and Madelaine Boyd and Anna-Luisa Brakman and Greg Brockman and Tim Brooks and Miles Brundage and Kevin Button and Trevor Cai and Rosie Campbell and Andrew Cann and Brittany Carey and Chelsea Carlson and Rory Carmichael and Brooke Chan and Che Chang and Fotis Chantzis and Derek Chen and Sully Chen and Ruby Chen and Jason Chen and Mark Chen and Ben Chess and Chester Cho and Casey Chu and Hyung Won Chung and Dave Cummings and Jeremiah Currier and Yunxing Dai and Cory Decareaux and Thomas Degry and Noah Deutsch and Damien Deville and Arka Dhar and David Dohan and Steve Dowling and Sheila Dunning and Adrien Ecoffet and Atty Eleti and Tyna Eloundou and David Farhi and Liam Fedus and Niko Felix and Simón Posada Fishman and Juston Forte and Isabella Fulford and Leo Gao and Elie Georges and Christian Gibson and Vik Goel and Tarun Gogineni and Gabriel Goh and Rapha Gontijo-Lopes and Jonathan Gordon and Morgan Grafstein and Scott Gray and Ryan Greene and Joshua Gross and Shixiang Shane Gu and Yufei Guo and Chris Hallacy and Jesse Han and Jeff Harris and Yuchen He and Mike Heaton and Johannes Heidecke and Chris Hesse and Alan Hickey and Wade Hickey and Peter Hoeschele and Brandon Houghton and Kenny Hsu and Shengli Hu and Xin Hu and Joost Huizinga and Shantanu Jain and Shawn Jain and Joanne Jang and Angela Jiang and Roger Jiang and Haozhun Jin and Denny Jin and Shino Jomoto and Billie Jonn and Heewoo Jun and Tomer Kaftan and Łukasz Kaiser and Ali Kamali and Ingmar Kanitscheider and Nitish Shirish Keskar and Tabarak Khan and Logan Kilpatrick and Jong Wook Kim and Christina Kim and Yongjik Kim and Jan Hendrik Kirchner and Jamie Kiros and Matt Knight and Daniel Kokotajlo and Łukasz Kondraciuk and Andrew Kondrich and Aris Konstantinidis and Kyle Kosic and Gretchen Krueger and Vishal Kuo and Michael Lampe and Ikai Lan and Teddy Lee and Jan Leike and Jade Leung and Daniel Levy and Chak Ming Li and Rachel Lim and Molly Lin and Stephanie Lin and Mateusz Litwin and Theresa Lopez and Ryan Lowe and Patricia Lue and Anna Makanju and Kim Malfacini and Sam Manning and Todor Markov and Yaniv Markovski and Bianca Martin and Katie Mayer and Andrew Mayne and Bob McGrew and Scott Mayer McKinney and Christine McLeavey and Paul McMillan and Jake McNeil and David Medina and Aalok Mehta and Jacob Menick and Luke Metz and Andrey Mishchenko and Pamela Mishkin and Vinnie Monaco and Evan Morikawa and Daniel Mossing and Tong Mu and Mira Murati and Oleg Murk and David Mély and Ashvin Nair and Reiichiro Nakano and Rajeev Nayak and Arvind Neelakantan and Richard Ngo and Hyeonwoo Noh and Long Ouyang and Cullen O'Keefe and Jakub Pachocki and Alex Paino and Joe Palermo and Ashley Pantuliano and Giambattista Parascandolo and Joel Parish and Emy Parparita and Alex Passos and Mikhail Pavlov and Andrew Peng and Adam Perelman and Filipe de Avila Belbute Peres and Michael Petrov and Henrique Ponde de Oliveira Pinto and Michael and Pokorny and Michelle Pokrass and Vitchyr H. Pong and Tolly Powell and Alethea Power and Boris Power and Elizabeth Proehl and Raul Puri and Alec Radford and Jack Rae and Aditya Ramesh and Cameron Raymond and Francis Real and Kendra Rimbach and Carl Ross and Bob Rotsted and Henri Roussez and Nick Ryder and Mario Saltarelli and Ted Sanders and Shibani Santurkar and Girish Sastry and Heather Schmidt and David Schnurr and John Schulman and Daniel Selsam and Kyla Sheppard and Toki Sherbakov and Jessica Shieh and Sarah Shoker and Pranav Shyam and Szymon Sidor and Eric Sigler and Maddie Simens and Jordan Sitkin and Katarina Slama and Ian Sohl and Benjamin Sokolowsky and Yang Song and Natalie Staudacher and Felipe Petroski Such and Natalie Summers and Ilya Sutskever and Jie Tang and Nikolas Tezak and Madeleine B. Thompson and Phil Tillet and Amin Tootoonchian and Elizabeth Tseng and Preston Tuggle and Nick Turley and Jerry Tworek and Juan Felipe Cerón Uribe and Andrea Vallone and Arun Vijayvergiya and Chelsea Voss and Carroll Wainwright and Justin Jay Wang and Alvin Wang and Ben Wang and Jonathan Ward and Jason Wei and CJ Weinmann and Akila Welihinda and Peter Welinder and Jiayi Weng and Lilian Weng and Matt Wiethoff and Dave Willner and Clemens Winter and Samuel Wolrich and Hannah Wong and Lauren Workman and Sherwin Wu and Jeff Wu and Michael Wu and Kai Xiao and Tao Xu and Sarah Yoo and Kevin Yu and Qiming Yuan and Wojciech Zaremba and Rowan Zellers and Chong Zhang and Marvin Zhang and Shengjia Zhao and Tianhao Zheng and Juntang Zhuang and William Zhuk and Barret Zoph},
      year={2024},
      eprint={2303.08774},
      archivePrefix={arXiv},
      primaryClass={cs.CL},
      url={https://arxiv.org/abs/2303.08774}, 
}

@article{fcd,
author = {Preuer, Kristina and Renz, Philipp and Unterthiner, Thomas and Hochreiter, Sepp and Klambauer, G{\"u}nter},
title = {Fréchet ChemNet Distance: A Metric for Generative Models for Molecules in Drug Discovery},
journal = {Journal of Chemical Information and Modeling},
volume = {58},
number = {9},
pages = {1736-1741},
year = {2018},
doi = {10.1021/acs.jcim.8b00234},
    note ={PMID: 30118593},

URL = { 
    
        https://doi.org/10.1021/acs.jcim.8b00234
    
    

},
eprint = { 
    
        https://doi.org/10.1021/acs.jcim.8b00234
    
    

}

}

@misc{vonwerra2022trl,
  author = {Leandro von Werra and Younes Belkada and Lewis Tunstall and Edward Beeching and Tristan Thrush and Nathan Lambert and Shengyi Huang and Kashif Rasul and Quentin Gallouédec},
  title = {TRL: Transformer Reinforcement Learning},
  year = {2020},
  publisher = {GitHub},
  journal = {GitHub repository},
  howpublished = {\url{https://github.com/huggingface/trl}}
}

@inproceedings{kwon2023efficient,
  title={Efficient Memory Management for Large Language Model Serving with PagedAttention},
  author={Woosuk Kwon and Zhuohan Li and Siyuan Zhuang and Ying Sheng and Lianmin Zheng and Cody Hao Yu and Joseph E. Gonzalez and Hao Zhang and Ion Stoica},
  booktitle={Proceedings of the ACM SIGOPS 29th Symposium on Operating Systems Principles},
  year={2023}
}

@misc{hu2021loralowrankadaptationlarge,
      title={LoRA: Low-Rank Adaptation of Large Language Models}, 
      author={Edward J. Hu and Yelong Shen and Phillip Wallis and Zeyuan Allen-Zhu and Yuanzhi Li and Shean Wang and Lu Wang and Weizhu Chen},
      year={2021},
      eprint={2106.09685},
      archivePrefix={arXiv},
      primaryClass={cs.CL},
      url={https://arxiv.org/abs/2106.09685}, 
}

@misc{gu2025surveyllmasajudge,
      title={A Survey on LLM-as-a-Judge}, 
      author={Jiawei Gu and Xuhui Jiang and Zhichao Shi and Hexiang Tan and Xuehao Zhai and Chengjin Xu and Wei Li and Yinghan Shen and Shengjie Ma and Honghao Liu and Saizhuo Wang and Kun Zhang and Yuanzhuo Wang and Wen Gao and Lionel Ni and Jian Guo},
      year={2025},
      eprint={2411.15594},
      archivePrefix={arXiv},
      primaryClass={cs.CL},
      url={https://arxiv.org/abs/2411.15594}, 
}

@article{tetko2020state,
  title={State-of-the-art augmented NLP transformer models for direct and single-step retrosynthesis},
  author={Tetko, Igor V and Karpov, Pavel and Van Deursen, Ruud and Godin, Guillaume},
  journal={Nature communications},
  volume={11},
  number={1},
  pages={5575},
  year={2020},
  publisher={Nature Publishing Group UK London}
}

@misc{liu2024moleculargptopenlargelanguage,
      title={MolecularGPT: Open Large Language Model (LLM) for Few-Shot Molecular Property Prediction}, 
      author={Yuyan Liu and Sirui Ding and Sheng Zhou and Wenqi Fan and Qiaoyu Tan},
      year={2024},
      eprint={2406.12950},
      archivePrefix={arXiv},
      primaryClass={q-bio.QM},
      url={https://arxiv.org/abs/2406.12950}, 
}

@misc{liu2024reactxtunderstandingmolecularreactionship,
      title={ReactXT: Understanding Molecular "Reaction-ship" via Reaction-Contextualized Molecule-Text Pretraining}, 
      author={Zhiyuan Liu and Yaorui Shi and An Zhang and Sihang Li and Enzhi Zhang and Xiang Wang and Kenji Kawaguchi and Tat-Seng Chua},
      year={2024},
      eprint={2405.14225},
      archivePrefix={arXiv},
      primaryClass={q-bio.QM},
      url={https://arxiv.org/abs/2405.14225}, 
}

@article{zuo2025ttrl,
  title={Ttrl: Test-time reinforcement learning},
  author={Zuo, Yuxin and Zhang, Kaiyan and Sheng, Li and Qu, Shang and Cui, Ganqu and Zhu, Xuekai and Li, Haozhan and Zhang, Yuchen and Long, Xinwei and Hua, Ermo and others},
  journal={arXiv preprint arXiv:2504.16084},
  year={2025}
}

@misc{rtgrpo,
      title={Improving Low-Resource Machine Translation via Round-Trip Reinforcement Learning}, 
      author={Ahmed Attia and Alham Fikri Aji},
      year={2026},
      eprint={2601.12535},
      archivePrefix={arXiv},
      primaryClass={cs.CL},
      url={https://arxiv.org/abs/2601.12535}, 
}

@misc{rtmol,
      title={RTMol: Rethinking Molecule-text Alignment in a Round-trip View}, 
      author={Letian Chen and Runhan Shi and Gufeng Yu and Yang Yang},
      year={2025},
      eprint={2511.12135},
      archivePrefix={arXiv},
      primaryClass={cs.AI},
      url={https://arxiv.org/abs/2511.12135}, 
}

@misc{duallearning,
      title={Dual Supervised Learning}, 
      author={Yingce Xia and Tao Qin and Wei Chen and Jiang Bian and Nenghai Yu and Tie-Yan Liu},
      year={2017},
      eprint={1707.00415},
      archivePrefix={arXiv},
      primaryClass={cs.LG},
      url={https://arxiv.org/abs/1707.00415}, 
}

@misc{cycle-con,
      title={Language Models and Cycle Consistency for Self-Reflective Machine Translation}, 
      author={Jianqiao Wangni},
      year={2024},
      eprint={2411.02791},
      archivePrefix={arXiv},
      primaryClass={cs.CL},
      url={https://arxiv.org/abs/2411.02791}, 
}

@misc{reactxt,
      title={ReactXT: Understanding Molecular "Reaction-ship" via Reaction-Contextualized Molecule-Text Pretraining},
      author={Zhiyuan Liu and Yaorui Shi and An Zhang and Sihang Li and Enzhi Zhang and Xiang Wang and Kenji Kawaguchi and Tat-Seng Chua},
      year={2024},
      eprint={2405.14225},
      archivePrefix={arXiv},
      primaryClass={q-bio.QM},
      url={https://arxiv.org/abs/2405.14225},
}

@misc{liu2025trustverifyselfverificationapproach,
      title={Trust, But Verify: A Self-Verification Approach to Reinforcement Learning with Verifiable Rewards},
      author={Xiaoyuan Liu and Tian Liang and Zhiwei He and Jiahao Xu and Wenxuan Wang and Pinjia He and Zhaopeng Tu and Haitao Mi and Dong Yu},
      year={2025},
      eprint={2505.13445},
      archivePrefix={arXiv},
      primaryClass={cs.AI},
      url={https://arxiv.org/abs/2505.13445},
}

@misc{yu2025rlprextrapolatingrlvrgeneral,
      title={RLPR: Extrapolating RLVR to General Domains without Verifiers},
      author={Tianyu Yu and Bo Ji and Shouli Wang and Shu Yao and Zefan Wang and Ganqu Cui and Lifan Yuan and Ning Ding and Yuan Yao and Zhiyuan Liu and Maosong Sun and Tat-Seng Chua},
      year={2025},
      eprint={2506.18254},
      archivePrefix={arXiv},
      primaryClass={cs.LG},
      url={https://arxiv.org/abs/2506.18254},
}

@misc{zhang2025freelunchrethinkinginternal,
      title={No Free Lunch: Rethinking Internal Feedback for LLM Reasoning},
      author={Yanzhi Zhang and Zhaoxi Zhang and Haoxiang Guan and Yilin Cheng and Yitong Duan and Chen Wang and Yue Wang and Shuxin Zheng and Jiyan He},
      year={2025},
      eprint={2506.17219},
      archivePrefix={arXiv},
      primaryClass={cs.LG},
      url={https://arxiv.org/abs/2506.17219},
}

@article{lindsey2025biology,
  author={Lindsey, Jack and Gurnee, Wes and Ameisen, Emmanuel and Chen, Brian and Pearce, Adam and Turner, Nicholas L. and Citro, Craig and Abrahams, David and Carter, Shan and Hosmer, Basil and Marcus, Jonathan and Sklar, Michael and Templeton, Adly and Bricken, Trenton and McDougall, Callum and Cunningham, Hoagy and Henighan, Thomas and Jermyn, Adam and Jones, Andy and Persic, Andrew and Qi, Zhenyi and Thompson, T. Ben and Zimmerman, Sam and Rivoire, Kelley and Conerly, Thomas and Olah, Chris and Batson, Joshua},
  title={On the Biology of a Large Language Model},
  journal={Transformer Circuits Thread},
  year={2025},
  url={https://transformer-circuits.pub/2025/attribution-graphs/biology.html}
}

@misc{liao2025lostbeginningreasoning,
      title={Lost at the Beginning of Reasoning},
      author={Baohao Liao and Xinyi Chen and Sara Rajaee and Yuhui Xu and Christian Herold and Anders Søgaard and Maarten de Rijke and Christof Monz},
      year={2025},
      eprint={2506.22058},
      archivePrefix={arXiv},
      primaryClass={cs.CL},
      url={https://arxiv.org/abs/2506.22058},
}
\bibliographystyle{colm2026_conference}

\newpage
\appendix
\section*{Appendix}
\section{Dataset Details}
\label{app:data}
This section provides a detailed description about each dataset used in the experiments. Examples of each dataset and task can be found in Table~\ref{tab:data_exp}.
\begin{table}[h] 
\centering 
\caption{Examples of data and tasks.} 
\small
\label{tab:data_exp} 
\resizebox{1.0\textwidth}{!}{
\begin{tabular}{p{2cm}p{2.8cm}p{6cm}p{6cm}} 
\hline
Dataset & Task & Input & Output \\
\hline
CHEBI-20 & Molecule Captioning & \seqsplit{COc1cccc2[nH]cc(C/C(=N/OS(=O)(=O)[O-])S[C@@H]3O[C@H](CO)[C@@H](O)[C@H](O)[C@H]3O)c12} & The molecule is an indolylmethylglucosinolate that is the conjugate base of 4-methoxyglucobrassicin, obtained by deprotonation of the sulfo group. It is a conjugate base of a 4-methoxyglucobrassicin. \\
\hline
CHEBI-20 & Text-based Molecule Generation & The molecule is a 1,3-thiazolium cation that is 1,3-thiazol-3-ium substituted by a methyl group at position 4, a (4-amino-2-methylpyrimidin-5-yl)methyl group at position 3, a 3-carboxy-1-hydroxypropyl at position 2 and a 2-{[hydroxy(phosphonooxy)phosphoryl]oxy}ethyl group at position 5. It has a role as a human metabolite and a mouse metabolite. &  \seqsplit{Cc1ncc(C[n+]2c(C(O)CCC(=O)O)sc(CCOP(=O)(O)OP(=O)(O)O)c2C)c(=N)[nH]1}\\
\hline
LM-24 & Molecule Captioning & \seqsplit{CCCCCCCCCCCCCCCCCCCCCC(=O)OC[C@H](COC(=O)CCCCCCCCCCCCCCCCCCCC)OC(=O)CCCCCCCCCC(C)C} & The molecule is a energy storage, nutrient, membrane stabilizer that impacts cardiovascular disease, metabolic syndrome, and pancreatitis. The molecule is a member of the thyroxine treatment class and affects both atherosclerosis and cancer. The molecule is a fat storage, a energy source, and a inflammatory, and it impacts obesity. \\
\hline
LM-24 & Text-based Molecule Generation & When heated to decomposition it emits acrid smoke and irritating fumes. The molecule has both a Bitter and unpleasant taste and a Pleasant odor. & \seqsplit{COc1ccc(C(C)=O)cc1} \\
\hline
USPTO-Mixed & Reaction Prediction & \seqsplit{CC\#N.CCN(CC)CC.COC(=O)C1(OC)CC(C)C(=O)C=C1O.Cc1nc(C(F)(F)F)ccc1C(=O)Cl.N\#[C][K]} & \seqsplit{COC(=O)C1(OC)CC(C)C(=O)C(C(=O)c2ccc(C(F)(F)F)nc2C)=C1O} \\
\hline
USPTO-50K & Retrosynthesis & \seqsplit{CNC(=O)c1ccc2c(c1)CC(=O)N2C1CCN(CC(=O)N2C[C@H]3CCC[C@H]3C2)CC1} &  \seqsplit{CNC(=O)c1ccc2c(c1)CC(=O)N2C1CCNCC1.O=C(CCl)N1C[C@H]2CCC[C@H]2C1}\\
\hline
\end{tabular}}
\end{table}

\textbf{CHEBI-20 Dataset:} The ChEBI-20 dataset, as prepared for the paper~\citet{edwards-etal-2021-text2mol}, is a cross-modal retrieval benchmark containing approximately 33,000 molecule-description pairs sourced from the ChEBI and PubChem databases. The core task is to retrieve a specific molecule from a large corpus using only its natural language description as a query. This requires a model to learn a shared semantic space bridging textual descriptions and molecular graph structures. Unlike versions used for role classification, this dataset is specifically designed to test a model's ability to understand fine-grained, descriptive text for direct molecule retrieval.

\textbf{Language-Plus-Molecule-24 Dataset:} The Language-Plus-Molecule-24 (LPM-24) dataset is a large-scale, multimodal dataset designed for text-molecule tasks, particularly cross-modal retrieval and molecule captioning~\citep{edwards-etal-2024-l}. It was created by mining US patent documents to extract pairs of molecule structures and their corresponding textual descriptions. LPM-24 is significantly larger and more diverse in its language complexity than previous benchmarks like ChEBI-20. The descriptions are often highly technical, focusing on the molecule's synthesis, properties, or application. This makes it a challenging benchmark for training and evaluating models on their ability to comprehend and align complex, domain-specific language with intricate molecular structures.

\textbf{USPTO-50K Dataset:} The USPTO-50K~\citep{upsto50} dataset for retrosynthesis is a benchmark derived from the standard USPTO-50k reaction prediction dataset. The goal is to predict a set of plausible reactants (precursors) that could synthesize a given product (target molecule). Models are trained to learn the logic of chemical disconnection and identify the key bonds to break in the target molecule. This dataset is fundamental for training and evaluating machine learning models, especially graph-based and transformer models, on their ability to perform single-step retrosynthesis, which is a critical component of computer-aided synthesis planning.

\textbf{USPTO-Mixed Dataset:} USPTO-Mixed~\citep{jin2017predicting} data is also derived from USPTO patent. Its corresponding task is to perform forward reaction prediction. It mixed the reactants and reagent to make the task even more challenging for downstream model to identify key reaction and generate correct products.
\section{More Related Works}
\label{app:more_rwork}

\textbf{Self-improving and Reinforcement Learning:} LLM self-improving has been a popular topic in various domains. Mostly in math and coding generation, self-improving techniques have achieved considerable success due to tools available for verifiable reward generation in these domains~\citep {gao2025survey,ma2025s,huang2025r,guanrstar}. Another line of work also facilitates unsupervised/self-supervised self-improving without ground-truth label~\citep{huang2023large,agarwal2025unreasonable,shao2025spurious,zuo2025ttrl}. These works strive to increase a model's confidence during output, while RTRL intends to check whether the output is consistent with the model's knowledge, providing a new lens into self-improving, particularly in the chemistry domain.
\section{Discussion on Primary RTRL Use Cases}
\label{app:rtrl_usecase}
The RTRL framework is highly adaptable, offering distinct advantages across various scenarios of data availability. Here we recommend use cases of RTRL. 

RTRL \textbf{enhances a single mapping}, $f: \mathcal{X} \rightarrow \mathcal{Y}$, via enforcing RTC with minimal data from the target domain, $\mathcal{Y}$. This is critical even in fields beyond chemistry, for example, in software engineering, migrating legacy codebases ($\mathcal{X}$) with limited examples of migrated code ($\mathcal{Y}$). By forcing the generator to produce outputs that its own inverse function can recognize, RTRL compels a deeper, more robust understanding of the forward task, even without ground-truth labels.

In our \textbf{iterative training scheme}, the forward and reverse models can co-evolve in an iterative manner. A key strength of this process is that the datasets for the forward and reverse tasks do not need to be paired. This allows the framework to leverage two independent, potentially disjoint collections of data to facilitate mutual improvement. This dramatically increases the flexibility of data collection. 

In a more traditional \textbf{supervised} setting where paired data $(X, Y)$ is available, RTRL serves as a strong regularizer. While standard supervised fine-tuning optimizes for accuracy on the primary task, RTRL introduces an additional objective that enforces logical consistency and reliability, which reduces overfitting and encourages the model to learn more robust and generalizable representations.

Finally, RTRL facilitates a \textbf{self-play} paradigm where the target dataset, $Y$, is generated \textbf{synthetically} by the model's own forward function from a seed dataset $X$. This removes the dependency on label collection. The RL process, with its inherent exploration, enables the model to probe its own understanding, search the problem domain with synthetic data, and reinforce consistent reasoning paths. The model can refine and uncover latent knowledge entirely without external supervision.

\paragraph{Generalizability Beyond Chemistry.} While this paper validates RTRL within computational chemistry, the framework's core prerequisites are domain-agnostic: (1) the existence of bidirectional mappings between representations, and (2) a base model with initial competence in both directions. Crucially, as shown above, RTRL does not require the training data to be paired---unpaired corpora from each domain suffice. These conditions are met in several domains beyond chemistry. A particularly compelling application is \textbf{code migration}, where large corpora of legacy code (e.g., COBOL, Fortran) and modern code (e.g., Python, Rust) exist abundantly, yet paired migration examples are scarce and expensive to produce. A model with basic translation ability between the two can apply RTRL to self-improve: forward-migrated code that fails the round-trip test (back-migration does not recover the original logic) receives low reward, encouraging more faithful translations. Similarly, \textbf{low-resource machine translation}~\citep{rtgrpo} and \textbf{natural-language-to-code} generation~\citep{allamanis2024unsupervised} present bidirectional structures where monolingual data is abundant but parallel corpora are limited. We plan to validate RTRL in these settings in future work, and are explicit that our current empirical scope is chemistry.
\section{Experiment Details}
\label{app:exp}
This section provides detailed experimental setup for reproducibility. The code repository is available at \href{https://github.com/LechengKong/RTRL}{https://github.com/LechengKong/RTRL}.

\textbf{Tools and Framework:} All experiments are conducted on a 8 Nvidia A40 node. The pretrained checkpoints are loaded from Hugging Face. The codes are written in PyTorch. For RL training, we use GRPO~\citep{shao2024deepseekmath} from TRL~\citep{vonwerra2022trl} package. We use vLLM~\citep{kwon2023efficient} for both completion generation in the forward function and round-trip consistency evaluation in the backward function. For finetuning, we use Hugging Face's trainer.

\textbf{Training hyperparameters:} For all RL training, we use 20000 samples and train for one epoch. For each sample we generate 12 completions (this non-$2^n$ number is because we use 6 GPU for training, 1 GPU for completion generation, and 1 GPU for round-trip consistency evaluation). We use a per device batch size 4, and gradient accumulation steps of 8. In total, the model witness 192 sample-completion-reward triplets in one optimization step. To sample completions we use Top-K of 40, Top-P of \{0.4,0.9\} subject to validation error, temperature of 0.9. The KL-divergence weight beta is set to either 0.04 or 0.08 depending on the validation error. For all finetuning, we use the full training set to finetune for 2 epochs, with a global batch size of 256. We use a learning rate of $2e-5$, weight decay factor of $0.001$, max gradient norm of $0.5$. We use a warm-up ratio of 0.03 to find the best learning rate, and then the learning rate follows cosine annealing schedule. All models are finetuned via Low-Rank Adaption (LoRA)~\citep{hu2021loralowrankadaptationlarge} with a rank of 32, alpha of 32, dropout of 0.05, and only the query and value projection matrix in the LLMs are optimized.

\textbf{Evaluation Metrics:} To comprehensively evaluate performance across textual and molecular modalities, we employ a diverse suite of metrics. For text generation tasks, we use standard NLP metrics (BLEU, ROUGE, and METEOR) to assess n-gram precision, recall, and semantic alignment. For molecule generation, we measure absolute accuracy using Exact Match—computed strictly after SMILES canonicalization—alongside sequence-level translation similarity metrics (BLEU and Levenshtein distance). To properly account for SMILES degeneracy and evaluate true chemical equivalence, we verify structural Validity via RDKit and compute Tanimoto similarities across three distinct fingerprinting methods (MACCS, RDKit, and Morgan) to ensure generated molecules share core topological and functional properties with the reference targets. Finally, we evaluate the global distributional quality and diversity of the generated molecule sets using Fréchet ChemNet Distance (FCD)~\citep{fcd}, computed directly against the target test sets to ensure statistical stability. We note that BLEU and Levenshtein distance are sequence-level surface metrics: on SMILES-output tasks they can be noisy, since chemically equivalent molecules may be written as syntactically different strings. We report them because they are standard in the chemical LLM literature (LlaSMol, ChemDFM, Mol-Instructions, and the LM-24 benchmark all report them for molecule tasks), enabling direct comparison, but they should be interpreted alongside the canonicalization-aware Exact Match and fingerprint-based similarity metrics, which measure true chemical equivalence.

\textbf{Text-based tasks:} In supervised RTRL, the round-trip consistency is combined with evaluated metrics of the generated outputs. The metric needs to be single scalar, so we simply add the BLEU-2, BLEU-4, METEOR, ROUGE-1, ROUGE-2, and ROUGE-L score. When combining this with the round-trip reward, we normalize it to a value between 0 and 1 add that to the round-trip reward.

\textbf{Molecule-based tasks:} In supervised RTRL, the metric is a sum of BLEU, MACCS similarity, RDKit similarity, and Morgan similarity. Other metrics are not included mostly because they are lower-the-better metric and has not upperbound.

For all self-supervised and iterative training setups, the unlabelled domains $\mathcal{X}$ and $\mathcal{Y}$ were sampled strictly from the established training splits of their respective datasets. All evaluation metrics and round-trip consistency tests were computed exclusively on the held-out test splits, ensuring no data leakage during the GRPO optimization phase.

\textbf{Iterative RTRL specification:} For iterative RTRL, we adopt a \emph{hard-switching} strategy: the generator is trained for one full epoch (one pass over the available data for that iteration), after which the forward and backward roles are swapped. We run 4 iterations in total. Each iteration is a standard, self-contained GRPO run reusing the hyperparameters above, so the only additional overhead relative to single-pass RTRL is swapping the frozen backward-model checkpoint between iterations, which is negligible compared to training. We stop at 4 iterations based on diminishing returns observed in validation metrics (Figure~\ref{fig:imp}); the current study uses the full available data to characterize how each direction benefits the other and when the improvement saturates, and we leave the study of optimal switching frequency to future work.

\textbf{Computational cost:} The cost scales linearly with the number of iterations. On our 8$\times$Nvidia A40 node, one epoch takes roughly 4 hours for reaction prediction (USPTO-Mixed) and roughly 5 hours for retrosynthesis (USPTO-50K); a 4-iteration iterative run therefore takes about 16 and 20 hours respectively, and about 36 hours for the combined bidirectional (reaction $+$ retrosynthesis) setting ($\sim$9 hours per iteration). Because each iteration is standard GRPO, RTRL scales with the same characteristics as any GRPO/PPO pipeline.

\subsection{Prompt Templates}\label{app:prompts}
We use the original task prompts released by the authors of the base chemical LLMs (LlaSMol and ChemDFM), and the full set is available in our code repository. As a representative example, the forward/backward prompt pair for reaction prediction / retrosynthesis is:

\textbf{Reaction Prediction (forward):} \texttt{Chemical reaction equations are typically expressed in the following form: reactant1.reactant2...>reagent1.reagent2...>product. \ldots{} Now we will provide you with an incomplete chemical reaction equation, where the missing part will be represented by ``\_\_\_''. \ldots{} Please only provide the missing part in your response, without any additional content. Incomplete equation: \{equa\}>>\_\_\_ Completion:}

\textbf{Retrosynthesis (backward):} \texttt{Chemical reaction equations are typically expressed in the following form: reactant1.reactant2...>>product. \ldots{} The missing parts could be one or more substances. \ldots{} Please only provide the missing part in your response, without any additional content. Incomplete equation: \_\_\_>>\{prod\} Completion:}

\subsection{Sensitivity to Backward Prompt}\label{app:promptsens}
Because RTRL's reward depends on the backward model, we test its sensitivity to the wording of the backward prompt. On CHEBI-20 molecule captioning (RTRL+LlaSMol), we compare the short prompt used in our main experiments (from the original LlaSMol authors), \texttt{``Give me a molecule that satisfies the conditions outlined in the description: \{desc\}''}, against a more elaborate expert-framed prompt, \texttt{``As an expert chemist, carefully analyze the following molecular description and generate the SMILES notation for a molecule that matches all described characteristics including functional groups, structural motifs, and physicochemical properties: \{desc\}''}.

\begin{table}[h]
\centering
\caption{Sensitivity of RTRL to the backward prompt formulation (CHEBI-20 molecule captioning, RTRL+LlaSMol).}\label{tab:promptsens}
\resizebox{0.9\linewidth}{!}{
\begin{tabular}{lcccccc}
\toprule
Backward Prompt & BLEU-2 $\uparrow$ & BLEU-4 $\uparrow$ & ROUGE-1 $\uparrow$ & ROUGE-2 $\uparrow$ & ROUGE-L $\uparrow$ & METEOR $\uparrow$ \\
\midrule
Base (no RTRL) & 0.386 & 0.275 & 0.490 & 0.306 & 0.423 & 0.419 \\
Short prompt & 0.434 & 0.290 & 0.535 & 0.324 & 0.449 & 0.440 \\
Complex prompt & 0.430 & 0.292 & 0.536 & 0.319 & 0.442 & 0.435 \\
\bottomrule
\end{tabular}}
\end{table}

As Table~\ref{tab:promptsens} shows, both prompts yield very similar improvements (within 1\% across all metrics) and both substantially outperform the base model. This indicates that the round-trip reward is driven by the model's underlying bidirectional understanding rather than surface-level prompt engineering: as long as the backward prompt conveys the correct task semantics, its specific wording has minimal impact on training outcomes.
\section{More Experimental Results}
\label{app:moreexp}
\begin{table}[h]
\centering
\caption{RTRL applied to different base models.}\label{tab:moremd}

\resizebox{1.0\textwidth}{!}{
\begin{tabular}{lcccccccc}
\toprule
Task / Model & BLEU $\uparrow$ & Lev. $\downarrow$ & Exact Match $\uparrow$ & MACCS SIM. $\uparrow$ & RDKit SIM. $\uparrow$ & Morgan SIM. $\uparrow$ & FCD $\downarrow$ & Validity $\uparrow$ \\
\midrule
\rowcolor{gray!25} \multicolumn{9}{l}{\textit{USPTO-50K Retrosynthesis}} \\
  Qwen-8B & 0.571 & 35.589 & 0.000 & 0.638 & 0.546 & 0.519 & 22.522 & 0.962 \\
  Qwen-8B+RTRL & \textbf{0.581} & \textbf{34.721} & \textbf{0.000} & \textbf{0.646} & \textbf{0.556} & \textbf{0.529} & \textbf{21.560} & \textbf{0.963} \\
\midrule
  Mol-Instructions & 0.370 & 33.138 & 0.202 & 0.779 & 0.641 & 0.601 & 8.264 & \textbf{1.000} \\
  Mol-Instructions+RTRL & \textbf{0.397} & \textbf{30.873} & \textbf{0.216} & \textbf{0.797} & \textbf{0.642} & \textbf{0.621} & \textbf{6.678} & \textbf{1.000} \\
\midrule
  LlaSMol & 0.818 & 16.027 & 0.349 & \textbf{0.845} & 0.773 & 0.734 & 0.579 & \textbf{0.999} \\
  LlaSMol+RTRL & \textbf{0.825} & \textbf{15.178} & \textbf{0.361} & 0.840 & \textbf{0.777} & \textbf{0.734} & \textbf{0.494} & 0.998 \\
\midrule
  ChemDFM & 0.583 & 36.762 & \textbf{0.171} & 0.796 & 0.738 & 0.650 & 22.347 & 0.988 \\
  ChemDFM+RTRL & \textbf{0.625} & \textbf{31.996} & 0.151 & \textbf{0.805} & \textbf{0.759} & \textbf{0.671} & \textbf{21.449} & \textbf{0.992} \\
\midrule
\rowcolor{gray!25} \multicolumn{9}{l}{\textit{USPTO-Mixed Reaction Prediction}} \\
  Qwen-8B & 0.231 & 62.644 & \textbf{0.005} & 0.368 & 0.271 & 0.251 & \textbf{21.436} & \textbf{0.942} \\
  Qwen-8B+RTRL & \textbf{0.255} & \textbf{61.644} & 0.003 & \textbf{0.386} & \textbf{0.297} & \textbf{0.286} & 21.537 & 0.932 \\
\midrule
  Mol-Instructions & 0.307 & 28.725 & 0.096 & 0.578 & 0.436 & 0.385 & 4.193 & \textbf{0.999} \\
  Mol-Instructions+RTRL & \textbf{0.338} & \textbf{24.539} & \textbf{0.148} & \textbf{0.600} & \textbf{0.442} & \textbf{0.402} & \textbf{2.381} & 0.988 \\
\midrule
  LlaSMol & 0.923 & \textbf{5.042} & 0.690 & 0.919 & 0.879 & 0.858 & \textbf{0.235} & 0.996 \\
  LlaSMol+RTRL & \textbf{0.938} & 5.057 & \textbf{0.704} & \textbf{0.931} & \textbf{0.895} & \textbf{0.866} & 0.892 & \textbf{0.996} \\
\midrule
  ChemDFM & 0.845 & 8.685 & 0.559 & 0.880 & 0.831 & 0.803 & 18.818 & \textbf{0.987} \\
  ChemDFM+RTRL & \textbf{0.857} & \textbf{7.934} & \textbf{0.601} & \textbf{0.895} & \textbf{0.851} & \textbf{0.823} & \textbf{0.171} & 0.985 \\
\bottomrule
\end{tabular}}
\end{table}

\begin{table}[h]
\centering
\caption{RTRL Additional Molecule Results.}\label{tab:more_mol}
\resizebox{1.0\textwidth}{!}{
\begin{tabular}{lcccccccc}
\toprule
Task / Model & BLEU $\uparrow$ & Lev. $\downarrow$ & Exact Match $\uparrow$ & MACCS SIM. $\uparrow$ & RDKit SIM. $\uparrow$ & Morgan SIM. $\uparrow$ & FCD $\downarrow$ & Validity $\uparrow$ \\
\midrule
\rowcolor{gray!25} \multicolumn{9}{l}{\textit{LM-24 Text-based Molecule Generation Supervised}} \\
  GPT (10-shot) & 0.595 & 44.179 & 0.000 & 0.608 & 0.493 & 0.404 & 17.065 & 0.925 \\
  Qwen-8B & 0.277 & 207.726 & 0.000 & 0.681 & 0.623 & 0.466 & 73.977 & 0.857 \\
  Mol-Instructions & 0.411 & 120.475 & 0.000 & 0.696 & 0.637 & 0.476 & 26.684 & \textbf{1.000} \\
  ChemDFM & 0.688 & 49.169 & \textbf{0.001} & 0.740 & 0.669 & 0.493 & 63.314 & 0.969 \\
  GRPO+ChemDFM & 0.689 & 49.118 & 0.000 & 0.739 & 0.673 & 0.492 & 62.019 & 0.976 \\ \midrule
\textbf{RTRL+ChemDFM} & \textbf{0.737} & \textbf{39.948} & \textbf{0.001} & \textbf{0.762} & \textbf{0.683} & \textbf{0.524} & \textbf{3.307} & 0.996 \\
\midrule
\rowcolor{gray!25} \multicolumn{9}{l}{\textit{CHEBI-20 Text-based Molecule Generation Supervised}} \\
  GPT (10-shot) & 0.529 & 42.030 & 0.119 & 0.632 & 0.517 & 0.454 & 21.707 & 0.731 \\
  Qwen-8B & 0.162 & 221.933 & 0.009 & 0.524 & 0.348 & 0.277 & 24.704 & 0.830 \\
  Mol-Instructions & 0.479 & 68.238 & 0.083 & 0.714 & 0.499 & 0.406 & 2.647 & \textbf{0.993} \\
  ChemDFM & 0.821 & 18.712 & 0.494 & 0.887 & 0.804 & 0.757 & 2.087 & 0.958 \\
  GRPO+ChemDFM & 0.811 & 19.690 & 0.504 & 0.872 & 0.793 & 0.758 & 2.034 & 0.968 \\ \midrule
\textbf{RTRL+ChemDFM} & \textbf{0.868} & \textbf{14.311} & \textbf{0.547} & \textbf{0.904} & \textbf{0.815} & \textbf{0.772} & \textbf{1.372} & 0.984 \\
\midrule
\rowcolor{gray!25} \multicolumn{9}{l}{\textit{USPTO-Mixed Reaction Prediction Supervised}} \\
  GPT (10-shot) & 0.685 & 16.269 & 0.418 & 0.790 & 0.746 & 0.702 & 6.785 & 0.881 \\
  Qwen-8B & 0.662 & 20.260 & 0.031 & 0.653 & 0.583 & 0.507 & 18.346 & 0.931 \\
  Mol-Instructions & 0.660 & 19.135 & 0.011 & 0.653 & 0.563 & 0.490 & 8.470 & \textbf{0.999} \\
  LlaSMol & 0.923 & \textbf{5.042} & 0.690 & 0.919 & 0.879 & 0.858 & 0.235 & 0.996 \\
  ChemDFM & 0.891 & 6.433 & 0.660 & 0.908 & 0.865 & 0.846 & 0.115 & 0.989 \\
  GRPO+ChemDFM & 0.892 & 6.451 & 0.650 & 0.909 & 0.866 & 0.845 & 0.125 & 0.994 \\ \midrule
  \textbf{RTRL+LlaSMol} & \textbf{0.938} & 5.057 & \textbf{0.704} & \textbf{0.931} & \textbf{0.895} & \textbf{0.866} & 0.892 & 0.996 \\
\textbf{RTRL+ChemDFM} & 0.896 & 6.358 & 0.665 & 0.909 & 0.868 & 0.848 & \textbf{0.115} & 0.985 \\
\midrule
\rowcolor{gray!25} \multicolumn{9}{l}{\textit{USPTO-50K Retrosynthesis}} \\
  GPT (0-shot) & 0.601 & 27.388 & 0.104 & 0.609 & 0.464 & 0.441 & 14.624 & 0.821 \\
  Qwen-8B & 0.571 & 35.589 & 0.000 & 0.638 & 0.546 & 0.519 & 22.522 & 0.962 \\
  Mol-Instructions & 0.370 & 33.138 & 0.202 & 0.779 & 0.641 & 0.601 & 8.264 & \textbf{1.000} \\
  LlaSMol & 0.824 & \textbf{15.659} & \textbf{0.352} & 0.846 & \textbf{0.775} & \textbf{0.737} & \textbf{0.560} & 0.999 \\
  ChemDFM & 0.583 & 36.762 & 0.171 & 0.796 & 0.738 & 0.650 & 22.347 & 0.988 \\ \midrule
  \textbf{RTRL+LlaSMol} & \textbf{0.830} & 16.591 & 0.344 & \textbf{0.853} & 0.760 & 0.736 & 0.660 & 0.999 \\
\textbf{RTRL+ChemDFM} & 0.625 & 31.996 & 0.151 & 0.805 & 0.759 & 0.671 & 21.449 & 0.992 \\
\bottomrule
\end{tabular}}
\end{table}
Table~\ref{tab:moremd} shows model performance when RTRL is applied to different base models. All experiment are in a self-supervised manner. We can see that RTRL brings improvement to all tested base LLM, showing its generalizability. This also indicates vast hidden/covered knowledge within pretrained LLM, and enforcing round-trip consistency can be intuitive way to utilize such knowledge and improve the model. This also shows that RTRL can potentially be applied to larger model whose ability to judge its own response is better. We leave this to future work.

Table~\ref{tab:more_mol} shows more molecule results. We observe consistent improvement of RTRL over base model on most tasks. When combined with RTRL, both LlaSMol and ChemDFM show improvement. The only exception is USPTO-50K for retrosynthesis. We observe RTRL is only comparable with the base model, we suspect that this is because LlaSMol has been exposed to some of the training data, and extra training causes overfitting.

Table~\ref{tab:mol-simp} and Table~\ref{tab:text-simp} compare RTRL performance against base model, three synthetic training version, and entropy minimization (EM). Synthetic Output means we have a domain data $X\in \mathcal{X}$, and task is mapping it to another domain $\mathcal{Y}$, we first ask the model to generate synthetic labels $Y^*$, and perform SFT on the dataset $(X, Y^*)$. Synthetic Input means the opposite: for the same $\mathcal{X}$ to $\mathcal{Y}$ mapping task, we have a domain data $Y\in\mathcal{Y}$, we first ask the model to generate synthetic $X^*$ and then SFT on $(X^*, Y)$. RTC combine the above, but we first use input to go through the round-trip, and filter non-RTC sample (non-exact-match molecule, and BLEU-4 < 0.3 for text). We additionally consider SSL (self-supervised learning), where we simply train the model with the same input in a autoregressive manner. For EM, we follow recent work in \citet{agarwal2025unreasonable} to use the negative entropy of the generation as the reward. This method increases model's generation confidence, which is also self-supervised. From the results, we can see that RTRL still outperforms the variants. More importantly, RTRL can consistently improve over the base model on almost every metric, while the three variants can cause degradation. In particular, SSL will sometimes cause performance collapse, which is potentially due to large gradient updates from SFT without a constraint on the update. On the other hand, we see that the three synthetic variants can bring overall performance improvement, which aligns with earlier finding that round-trip consistency examples can improve a model's ability~\citep{alberti2019synthetic}. Synthetic Output is underperforming, as it might strengthen the existing overfitted behavior, causing worse evaluation performance. Synthetic Input shows better results, as it is implicitly enforcing a Round-trip consistency, by training the model to agree with its backward function, and RTRL is a systematic and organized way to perform such alignment. Comparing EM and RTRL, we see that EM comes closer to RTRL compared to the synthetic methods especially in the molecule task. However, we still see large gap between EM and RTRL on the text-based task. We suspect that this is because RTRL not only boost the confidence, but also activates more hidden knowledge to self-validate such confidence, and hence performs better.

\begin{table}[h]
\centering
\caption{RTRL on molecule task compared to training with synthetic Round-trip examples.}\label{tab:mol-simp}
\resizebox{\linewidth}{!}{
\begin{tabular}{lcccccccc}
\toprule
Task / Model & BLEU $\uparrow$ & Lev. $\downarrow$ & Exact Match $\uparrow$ & MACCS SIM. $\uparrow$ & RDKit SIM. $\uparrow$ & Morgan SIM. $\uparrow$ & FCD $\downarrow$ & Validity $\uparrow$ \\
\midrule
\rowcolor{gray!25} \multicolumn{9}{l}{\textit{CHEBI-20}} \\
  BASE & 0.846 & 15.652 & 0.546 & 0.898 & 0.779 & 0.733 & 2.088 & 0.978 \\
  SSL & 0.694 & 35.160 & 0.379 & 0.721 & 0.683 & 0.706 & 17.919 & \textbf{0.994} \\
  Synthetic Output & 0.713 & 30.175 & 0.491 & 0.888 & 0.777 & 0.732 & 10.098 & 0.974 \\
  Synthetic Input & 0.820 & 17.879 & 0.548 & 0.899 & 0.778 & 0.733 & 2.112 & 0.980 \\
  RTC & 0.814 & 18.610 & 0.533 & 0.874 & 0.751 & 0.726 & 3.094 & 0.978 \\
  EM & 0.843 & 15.990 & 0.549 & 0.895 & 0.774 & 0.733 & \textbf{2.086} & 0.978 \\
  \textbf{RTRL} & \textbf{0.852} & \textbf{15.022} & \textbf{0.553} & \textbf{0.901} & \textbf{0.781} & \textbf{0.735} & 2.099 & 0.982 \\
\bottomrule
\end{tabular}}
\end{table}

\begin{table}[h]
\centering
\caption{RTRL on text task compared to training with synthetic Round-trip examples.}\label{tab:text-simp}
\resizebox{\linewidth}{!}{
\begin{tabular}{lcccccc}
\toprule
Task / Model & BLEU-2 $\uparrow$ & BLEU-4 $\uparrow$ & ROUGE-1 $\uparrow$ & ROUGE-2 $\uparrow$ & ROUGE-L $\uparrow$ & METEOR $\uparrow$ \\
\midrule
\rowcolor{gray!25} \multicolumn{7}{l}{\textit{CHEBI-20}} \\
  BASE & 0.286 & 0.244 & 0.406 & 0.312 & 0.378 & 0.345 \\
  SSL & 0.280 & 0.227 & 0.385 & 0.294 & 0.363 & 0.339 \\
  Synthetic Output & 0.371 & 0.320 & 0.475 & 0.371 & 0.439 & 0.418 \\
  Synthetic Input & 0.379 & 0.325 & 0.478 & 0.371 & 0.442 & 0.425 \\
  RTC & 0.389 & 0.348 & 0.481 & 0.373 & 0.453 & 0.439 \\
  EM & 0.390 & 0.333 & 0.486 & 0.376 & 0.448 & 0.431 \\
  \textbf{RTRL} & \textbf{0.447} & \textbf{0.380} & \textbf{0.529} & \textbf{0.406} & \textbf{0.483} & \textbf{0.481} \\
\bottomrule
\end{tabular}}
\end{table}

\begin{wraptable}{r}{5cm}
\vspace{-15pt}
    \centering
\caption{RTRL applied to classification tasks.}\label{tab:mol-cls}
\resizebox{\linewidth}{!}{
\begin{tabular}{lcc}
\toprule
Model & BBBP $\uparrow$ & SIDER $\uparrow$ \\
\midrule
Llasmol & 0.741 & 0.701 \\
Llasmol+SFT & 0.741 & 0.689 \\
\textbf{Llasmol+RTRL} & \textbf{0.757} & \textbf{0.713} \\
\bottomrule
\end{tabular}}
\end{wraptable}
We can additionally applied RTRL to classification tasks. Specifically, the forward function is now the classification task, such as predicting if a molecule will have a certain property. And then the backward function is asking the model to generate a molecule that will/will not have this property. This is essentially a conditional generation task that LLM is capable of. Given this observation, we conducted experiments on RTRL's ability in classification tasks. We first use SFT on the model, and then apply RTRL to this model. The results are in Table~\ref{tab:mol-cls}.
From the result we see, RTRL can actually improve the performance of the base chemical LLM on classification tasks. This shows a higher degree of flexibility of RTRL.

\section{Qualitative Examples}
\label{app:example}
This section discuss some qualitative examples comparing base models and RTRL-improved models.

\begin{table}[h] 
\centering 
\caption{Examples of molecule captioning outputs across different iteration in iterative RTRL.} 
\small
\label{tab:cap_imp} 
\resizebox{1.0\textwidth}{!}{
\begin{tabular}{p{3cm}p{3cm}p{3cm}p{3cm}p{3cm}} 
\hline
Molecule & Base Model Caption & Iteration 1 Caption & Iteration 2 Caption & Ground Truth\\
\hline
\raisebox{-1\height}{\includegraphics[width=\linewidth, trim={3cm 1cm 3cm 1cm}, clip]{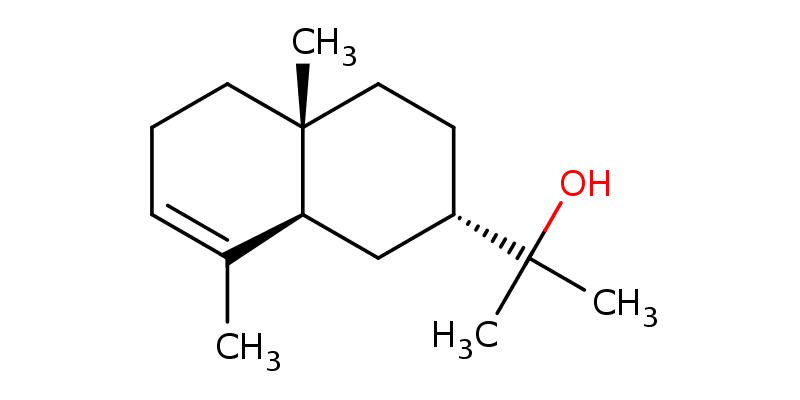}} \seqsplit{CC1=CCC[C@]2(C)CC[C@H](C(C)(C)O)C[C@@H]12} & The molecule is a natural product found in Hedychium spicatum with data available. & The molecule is a eudesmane sesquiterpenoid in which the eudesmane skeleton carries a hydroxy substituent at C-11 and has a double bond between C-3 and C-4. It has a role as a volatile oil component. & The molecule is a eudesmane sesquiterpenoid in which the eudesmane skeleton carries a hydroxy substituent at C-11 and has a double bond between C-3 and C-4. It has a role as a volatile oil component. It is a eudesmane sesquiterpenoid, a member of octahydronaphthalenes and a tertiary alcohol.
& The molecule is a eudesmane sesquiterpenoid in which the eudesmane skeleton carries a hydroxy substituent at C-11 and has a double bond between C-3 and C-4. It has a role as a plant metabolite and a volatile oil component. It is a eudesmane sesquiterpenoid, a member of octahydronaphthalenes and a tertiary alcohol. \\
\hline
\raisebox{-1\height}{\includegraphics[width=\linewidth, trim={3cm 1cm 3cm 1cm}, clip]{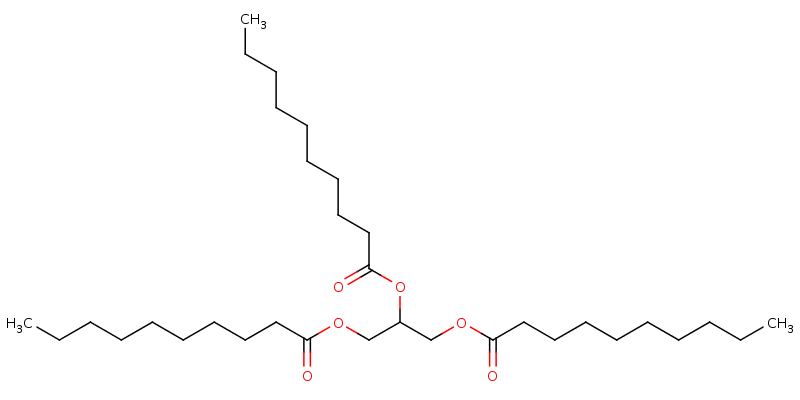}} \seqsplit{CCCCCCCCCC(=O)OCC(COC(=O)CCCCCCCCC)OC(=O)CCCCCCCCC} & The molecule is a triglyceride. & The molecule is a triglyceride that is glycerol in which all three hydroxy groups have been formally esterified with capric acid. It has a role as a Mycoplasma genitalium metabolite. It is a triglyceride, a caprate ester and a triacylglycerol 10:0. & The molecule is a triglyceride obtained by formal acylation of the three hydroxy groups of glycerol by decanoic acid. It has a role as a metabolite. It is a triglyceride and a decanoate ester.
& The molecule is a triglyceride obtained by formal acylation of the three hydroxy groups of glycerol by capric (decanoic) acid. It is a triglyceride and a decanoate ester. \\
\hline
\end{tabular}}
\end{table}

\begin{table}[h] 
\centering 
\caption{Examples comparing the outputs of RTRL-ChemDFM and ChemDFM on the reaction prediction task.} 
\small
\label{tab:react_pred} 
\resizebox{1.0\textwidth}{!}{
\begin{tabular}{p{4cm}p{4cm}p{4cm}p{4cm}} 
\hline
Reaction & Ground Truth & RTRL Prediction & Base Model Prediction \\
\hline
\raisebox{-1\height}{\includegraphics[width=\linewidth]{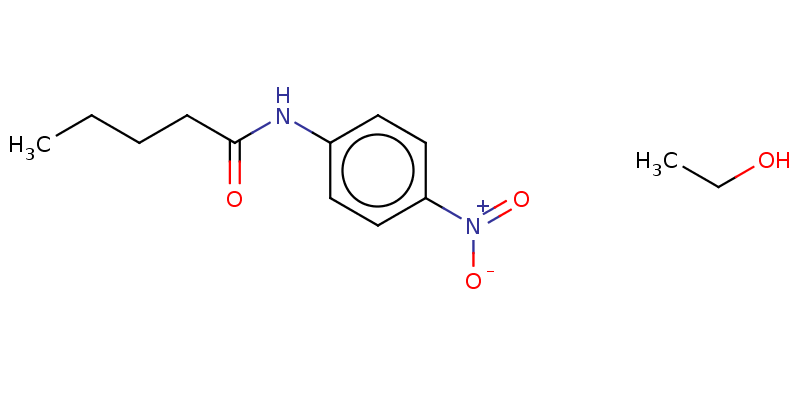}} \seqsplit{CCCCC(=O)Nc1ccc([N+](=O)[O-])cc1.CCO} & \raisebox{-1\height}{\includegraphics[width=\linewidth]{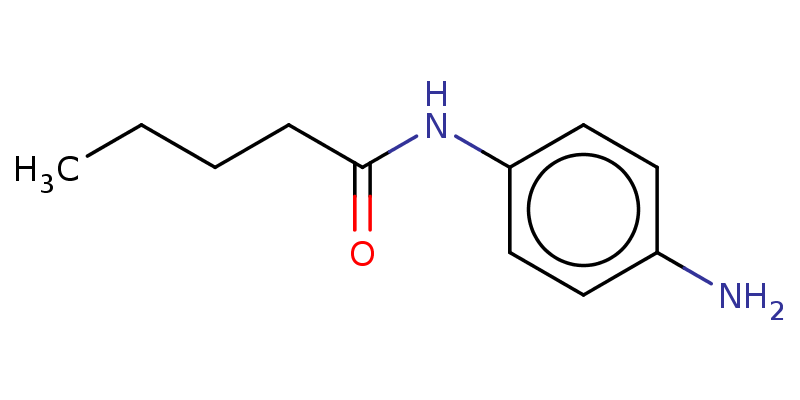}} \seqsplit{CCCCC(=O)Nc1ccc(N)cc1} & \raisebox{-1\height}{\includegraphics[width=\linewidth]{fig/mol_figs/mol_3.png}} \seqsplit{CCCCC(=O)Nc1ccc(N)cc1} & \raisebox{-1\height}{\includegraphics[width=\linewidth]{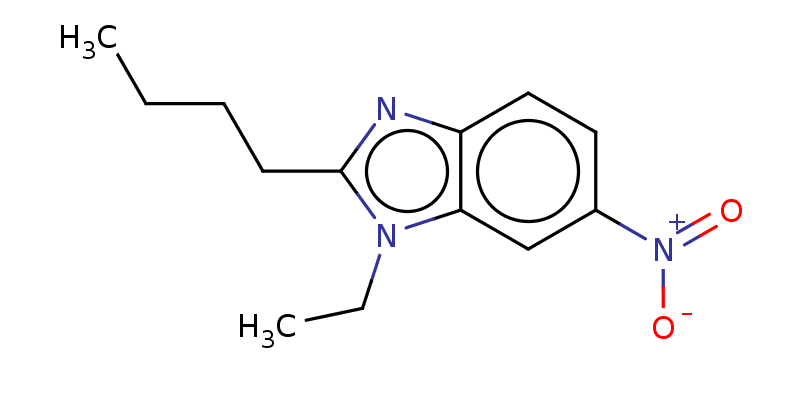}} \seqsplit{CCCCc1nc2ccc([N+](=O)[O-])cc2n1CC}
\\
\hline
\raisebox{-1\height}{\includegraphics[width=\linewidth]{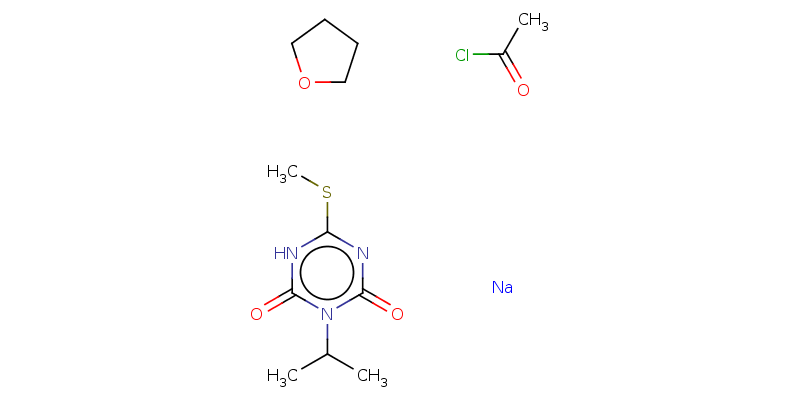}} \seqsplit{C1CCOC1.CC(=O)Cl.CSc1nc(=O)n(C(C)C)c(=O)[nH]1.[Na]} & \raisebox{-1\height}{\includegraphics[width=\linewidth]{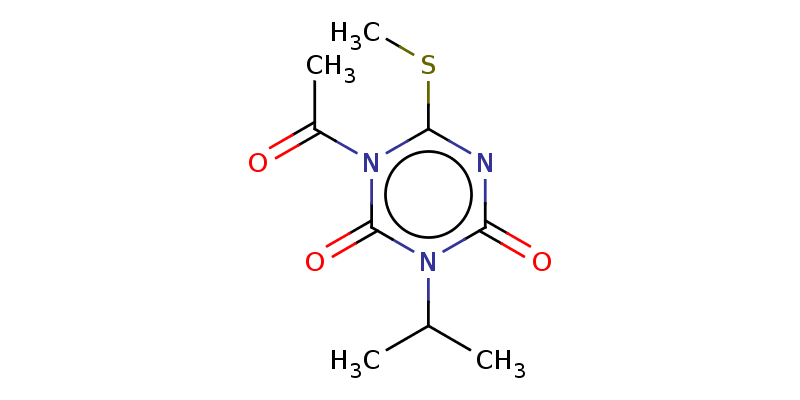}} \seqsplit{CSc1nc(=O)n(C(C)C)c(=O)n1C(C)=O} & \raisebox{-1\height}{\includegraphics[width=\linewidth]{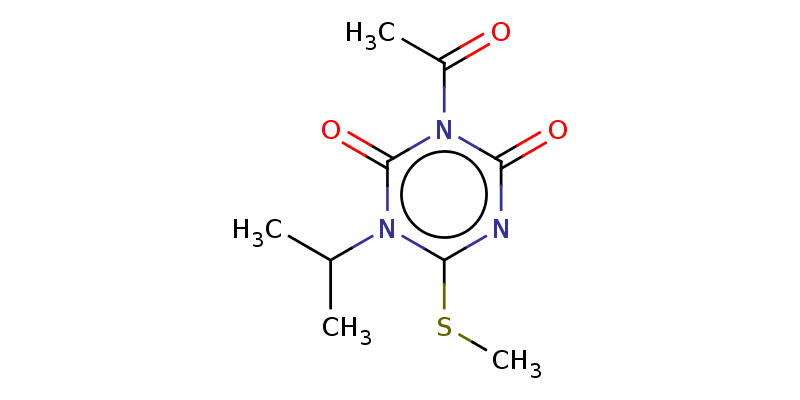}} \seqsplit{CC(=O)n1c(=O)nc(SC)n(C(C)C)c1=O} & \raisebox{-1\height}{\includegraphics[width=\linewidth]{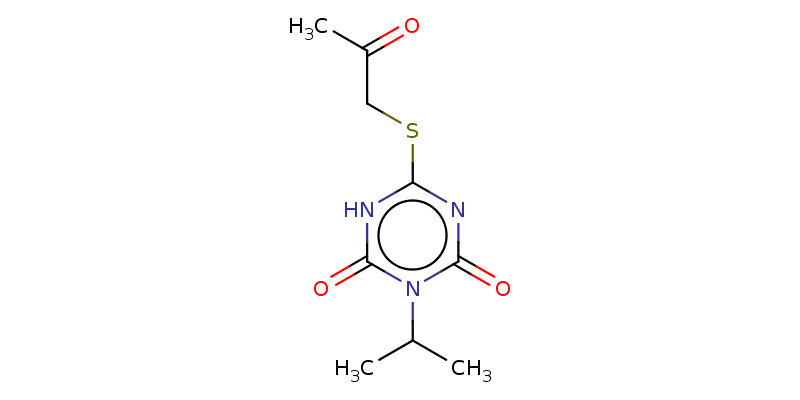}} \seqsplit{CC(=O)CSc1nc(=O)n(C(C)C)c(=O)[nH]1}
\\
\hline
\end{tabular}}
\end{table}

\begin{table}[h] 
\centering 
\caption{Examples comparing the outputs of RTRL-ChemDFM and ChemDFM on the retrosynthesis task.} 
\small
\label{tab:retro_pred} 
\resizebox{1.0\textwidth}{!}{
\begin{tabular}{p{4cm}p{4cm}p{4cm}p{4cm}} 
\hline
Product & Ground Truth & RTRL Prediction & Base Model Prediction \\
\hline
\raisebox{-1\height}{\includegraphics[width=\linewidth]{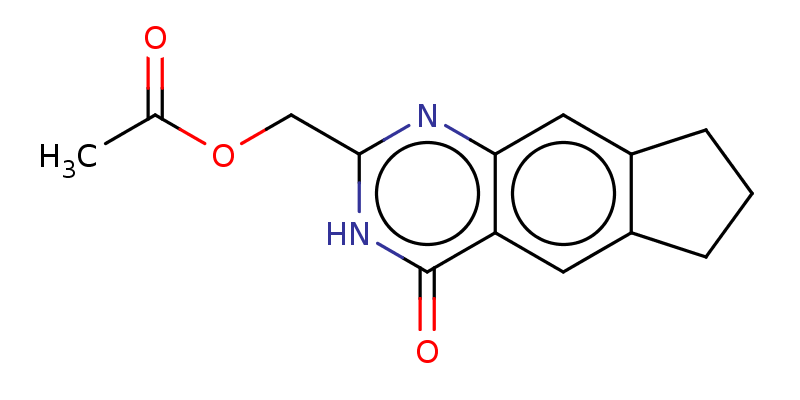}} \seqsplit{CC(=O)OCc1nc2cc3c(cc2c(=O)[nH]1)CCC3} & \raisebox{-1\height}{\includegraphics[width=\linewidth]{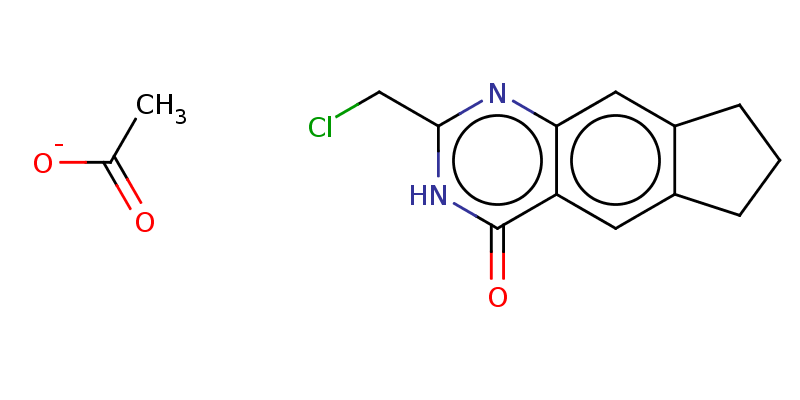}} \seqsplit{CC(=O)[O-].O=c1[nH]c(CCl)nc2cc3c(cc12)CCC3} & \raisebox{-1\height}{\includegraphics[width=\linewidth]{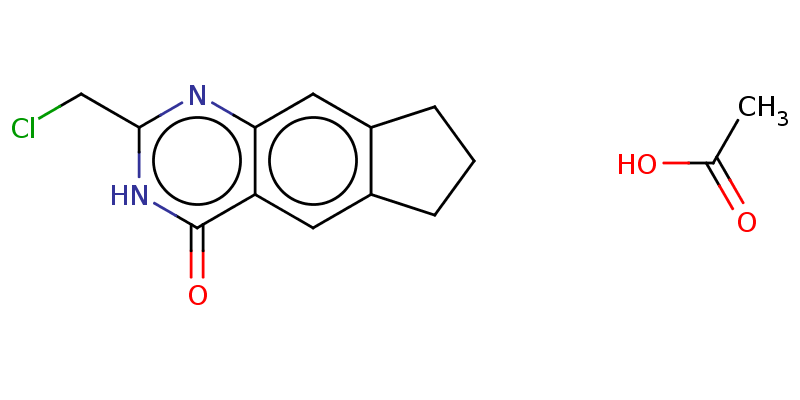}} \seqsplit{O=c1[nH]c(CCl)nc2cc3c(cc12)CCC3.CC(=O)O} & \raisebox{-1\height}{\includegraphics[width=\linewidth]{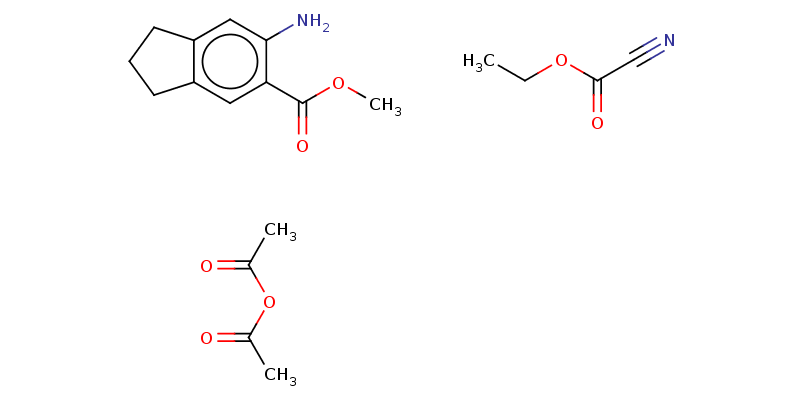}} \seqsplit{COC(=O)c1cc2c(cc1N)CCC2.CCOC(=O)C\#N.CC(=O)OC(C)=O}
\\
\hline
\raisebox{-1\height}{\includegraphics[width=\linewidth]{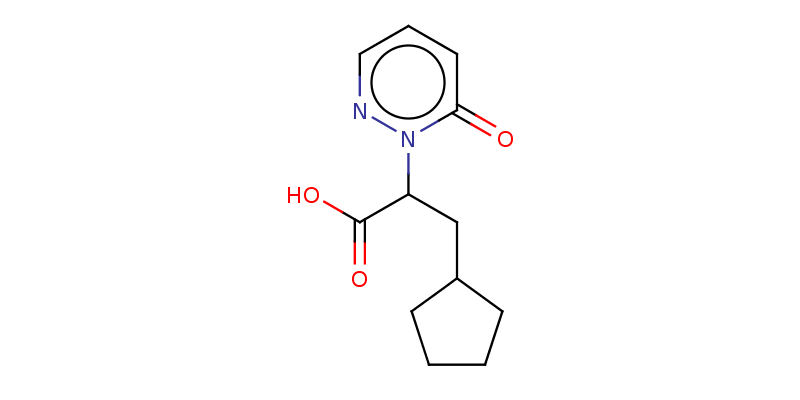}} \seqsplit{O=C(O)C(CC1CCCC1)n1ncccc1=O} & \raisebox{-1\height}{\includegraphics[width=\linewidth]{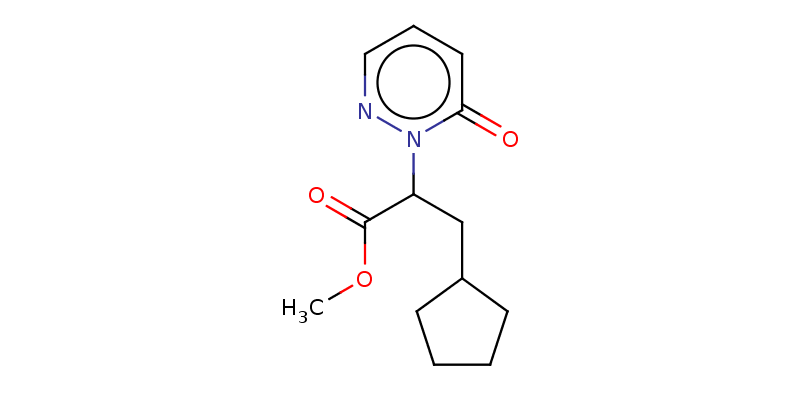}} \seqsplit{COC(=O)C(CC1CCCC1)n1ncccc1=O} & \raisebox{-1\height}{\includegraphics[width=\linewidth]{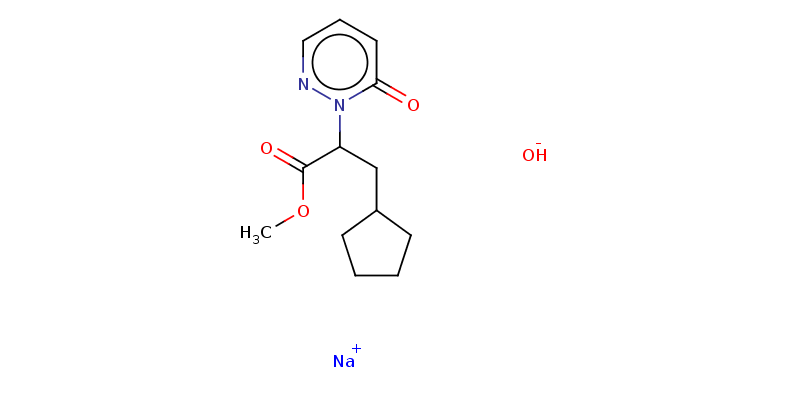}} \seqsplit{COC(=O)C(CC1CCCC1)n1ncccc1=O.[OH-].[Na+]} & \raisebox{-1\height}{\includegraphics[width=\linewidth]{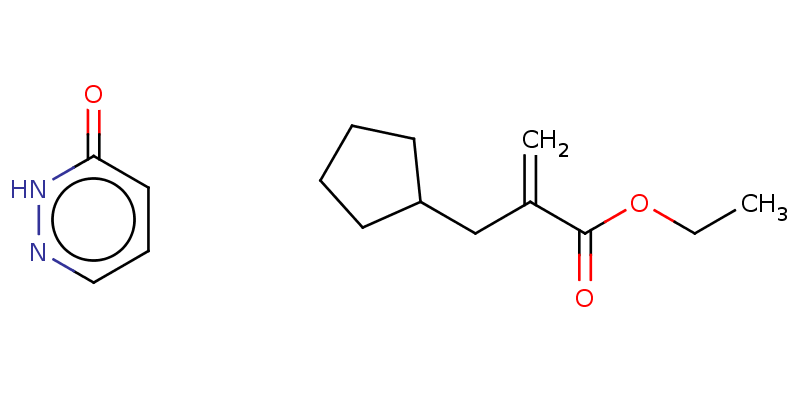}} \seqsplit{O=c1cccn[nH]1.C=C(CC1CCCC1)C(=O)OCC}
\\
\hline
\end{tabular}}
\end{table}
\textbf{Self-supervised RTRL:} Table~\ref{tab:react_pred} shows examples of reaction prediction. We compare ground truth with RTRL prediction and Base Model prediction. We can see that RTRL generally finds more accurate prediction and even exact matches. Table~\ref{tab:retro_pred} shows examples of retrosynthesis. Similarly, RTRL can generate more chemically correct examples. Meanwhile, as discussed earlier in the paper, we also see RTRL prefers to generate a complete reaction, like the example in the second row, where the main reactant is captured fully.

\textbf{Iterative RTRL:} Table~\ref{tab:cap_imp} shows generated examples when we use iterative RL to improve a model's molecule captioning ability. The two examples both show that the first iteration will teach the model to generate more specified and informative example, while the second iteration will further refine the model from the first iteration. However, we also notice interesting difference between the two example. In the first example, iteration 2 caption adds in missing details to iteration 1 caption. On the other hand, in the second example, iteration 2 caption removes erroneous description from iteration 1 caption. This shows that the model does not merely learn to generate longer responses especially in the iterative scenario, it can correct itself via round-trip consistency.
\section{More about GRPO}
\label{app:grpo}
The RTRL framework proposed in the paper can be extended to various RL method. In this paper, we use GRPO~\citep{shao2024deepseekmath} to instantiate the framework, and we introduce GRPO here for completeness.

GRPO is a RL algorithm that finetunes a policy LLM without a explicit value function. For a prompt $t$, the policy LLM generates a group of responses $G$, and the advantage for each generation in the group is computed as the normalized advantage over the average group performance:
\begin{equation}
    \hat{A}_i = \frac{r_i-mean(r_1,...,r_{|G|})}{std(r_1,...,r_{|G|})+\epsilon_{norm}} \quad \forall i\in G,
\end{equation}
where $\epsilon_{norm}$ is a small value for numerical stability and $r_i$ is the scaler reward for each sample, and the advantages are used as signal to guide the policy update. GRPO uses a clipped objective to optimize the model,
\begin{equation}
    \mathcal{L}(\theta)=-\frac{1}{|G|}\sum_i^{|G|}min(\frac{\pi_\theta(x_i)}{\pi_{\theta_{old}}(x_i)}\hat{A}_i, clip(\frac{\pi_\theta(x_i)}{\pi_{\theta_{old}}(x_i)},1-\epsilon,1+\epsilon)\hat{A}_i)+\beta KL(\pi_\theta(x_i), \pi_{\theta_{old}}(x_i)),
\end{equation}
where $x_i$ are the input sample, $\beta$ is hyperparameter controling the effect of the KL divergence term, $\pi_\theta$ is a policy/reference under the parameter $\theta$. This method belongs to the class of trust-region method, to contraint the magnitude of each upgrade. More details can be found in the original paper~\citep{shao2024deepseekmath}.

\section{More Discussion on Format Reward}
\label{app:format_reward}
As introduced in the main text, the format reward acts as a necessary structural guardrail, analogous to standard GRPO implementations. In standard GRPO for tasks like math or coding, rewards are often discrete (e.g., a score of 1 for a correct answer/format, and 0 otherwise). In RTRL, however, the primary objective is the continuous round-trip log-likelihood. Because this continuous reward is upper-bounded by 0 but lacks a strict lower bound, a simple binary format reward ($F \in \{0, 1\}$) cannot independently guarantee that format-incorrect outputs are penalized sufficiently.

To align with the GRPO principle that format correctness is a strict prerequisite, we dynamically scale the format reward by $\alpha = -m$, where $m$ is the empirical minimum round-trip log-likelihood observed over a preliminary set of training steps. This scaling guarantees that a structurally invalid answer ($F=0$) can never achieve a higher total reward than a format-correct answer ($F=1$), regardless of its likelihood score. Consequently, $\alpha$ does not require manual tuning; it is derived directly from the empirical reward distribution to enforce the structural constraint.

\begin{table}[h]
\centering
    \resizebox{0.9\textwidth}{!}{
\begin{tabular}{llcc}
\toprule
\textbf{Task} & \textbf{Modality} & \textbf{With Format Reward} & \textbf{Without Format Reward} \\
\midrule
Molecule Generation & Cross-modal & 0.553 & 0.550 \\
Reaction Prediction & Intra-modal & 0.601 & 0.597 \\
Retrosynthesis & Intra-modal & 0.323 & 0.157 \\
\bottomrule
\end{tabular}}
    \caption{Exact match results with or without format reward using ChemDFM base model.}
    \label{tab:abla_format}
\end{table}

An inherent empirical benefit of enforcing this standard format guardrail is its ability to prevent trivial identity-mapping. In our experiment, we observed this behavior predominantly in retrosynthesis in intra-modal tasks. Because both the input and output domains consist of SMILES strings, the forward model could theoretically maximize the round-trip reward by simply echoing the input reactants. In this scenario, the format reward elegantly eliminates the shortcut by utilizing cheminformatics tools (e.g., RDKit) to explicitly penalize outputs that do not parse as distinct, valid multiple chemicals.

Conversely, for cross-modal tasks (e.g., translating between molecules and text in CHEBI-20), the model naturally resists identity-mapping. Because advanced LLMs possess strong instruction-following capabilities, feeding a generated SMILES string back into the backward judge alongside a text-generation prompt directly violates the conditioning. The judge model recognizes the modality mismatch, resulting in a naturally low log-likelihood. Table~\ref{tab:abla_format} validates this. Thus, the format reward serves as a surgical safeguard: it remains largely dormant when inherent instruction-following suffices, but actively neutralizes mode collapse in susceptible intra-modal tasks.

\textbf{Future Directions.} While this lightweight formatting constraint effectively maintains structural integrity across our evaluated chemical tasks, exploring advanced techniques to enforce non-trivial transformations remains an interesting direction for the broader RL community. Future work could investigate min-max optimization frameworks that explicitly penalize high textual similarity between the input and output, or employ secondary LLMs as specialized judges to verify the semantic depth of the bidirectional transformation.
\section{More Discussion on the Limitations and Future Work}
\label{app:limitation}

To rigorously evaluate Round-Trip Consistency (RTC) as a trainable objective, our experimental design prioritized controlled environments and stable optimization. Consequently, we deferred several broader generative modeling challenges to future work. Below, we detail the rationale behind these scope boundaries and outline pathways for subsequent research.

\subsection{The Transition to Fully Synthetic Regimes}

Currently, RTRL relies on an external, unlabelled seed dataset from the target domain (e.g., a corpus of known SMILES strings) to initiate the self-improvement loop. While we demonstrated that RTRL successfully operates without paired labels, it does not yet operate in a completely zero-data, fully synthetic regime where the model must autonomously propose the initial inputs.

In computational chemistry, unlabelled molecules are abundant, making our current data-efficient paradigm highly practical. Expanding RTRL to bootstrap entirely from synthetic, model-generated distributions introduces the risk of compounding hallucinations, where the model optimizes for consistency over chemically invalid structures. Managing this distribution shift requires distinct out-of-distribution detection mechanisms. By anchoring our current framework to empirical seed data, we ensured that the observed performance gains were driven by the RTC objective rather than the complexities of managing open-ended synthetic generation.

\subsection{Precision Versus Generation Diversity}

The concern of reduced diversity has two distinct sources, which we address separately.

\textbf{SMILES non-uniqueness.} The same molecule can be written as many syntactically different but chemically equivalent SMILES strings. This is largely mitigated in our setting. LlaSMol is trained exclusively on canonicalized SMILES, making it strongly predisposed to generate canonical forms; since both the generator and the backward judge operate in canonical space, the round-trip evaluation naturally aligns with---rather than fights against---the model's existing behavior. For ChemDFM, which is trained on more diverse non-canonical SMILES, this is more relevant, but all our evaluation metrics (Exact Match, fingerprint similarities) are computed after SMILES canonicalization (Appendix~\ref{app:exp}), so chemically equivalent molecules are always recognized as correct regardless of their surface representation. Empirically, we did not observe diversity collapse arising from this source.

\textbf{Genuinely one-to-many tasks.} A more fundamental trade-off arises when a single input admits multiple distinct valid outputs---such as multiple retrosynthetic pathways for a product, or conditional molecule generation for a single label. Because RTRL uses a likelihood-based reward, the policy is naturally guided toward the most reliable, high-probability transformation that the backward model can seamlessly invert. For \emph{retrosynthesis}, we observe that the model still produces diverse outputs: because both the input molecules and their valid decompositions are diverse and we train only one epoch per iteration, the model learns general knowledge rather than fixating on point solutions. The more critical scenario is when the input set is narrow, e.g., \emph{conditional generation} where many molecules map to the same coarse label; here the policy can collapse onto a few ``safe'' molecules. This behavior is effective for maximizing exact match, validity, and structural coherence, but suppresses lower-probability, diverse routes---a limitation shared by most self-evolving generative techniques. Future implementations can address this by integrating established diversity-promoting techniques---entropy bonuses, diverse beam search decoding, GFlowNet-inspired objectives, or an explicit diversity reward computed via RDKit fingerprint dissimilarity---without altering the fundamental RTRL architecture.

\subsection{Stationary Reward Landscapes and Judge Co-evolution}

Throughout our implementation, the parameters of the backward judge model ($\phi$) are kept strictly frozen while the forward policy model ($\theta$) is updated via GRPO. Reinforcement learning is notoriously sensitive to non-stationary environments; a shifting reward signal can make credit assignment ambiguous and destabilize policy convergence.

Alternative architectures, such as dynamically updating the judge model, introduces a moving target for the generator. By freezing the judge, we effectively utilized it as a strict control variable. This design choice isolates the training dynamics, proving unequivocally that the model's performance improvements stem directly from adapting to the RTC objective, rather than from entangled co-training effects. Exploring the stable co-evolution of the forward and backward policies via EMA or alternating updates remains a highly promising direction for scaling the RTRL framework.

\subsection{Compatibility with Reasoning Models}

Our experiments use instruction-tuned base models that emit answers directly, and a natural question is whether RTRL extends to reasoning-first models that produce chain-of-thought before the final answer. We note two points. First, RTRL's formulation does not restrict the generator: GRPO treats the entire rollout (reasoning $+$ answer) as a single sequence receiving one reward, exactly as in standard reasoning RL, so the forward model is free to produce intermediate reasoning. Second, the backward judge remains meaningful even with reasoning outputs. Current instruct models (including LlaSMol) already interleave explanatory text with the final SMILES/reaction output, and our direct-concatenation forward pass already yields strong results. Moreover, recent work suggests that LLMs form a rough representation of the final answer before generating the reasoning chain~\citep{lindsey2025biology,liao2025lostbeginningreasoning}, so the backward likelihood $p(x \mid \text{answer}, \text{backward prompt})$ remains a valid signal: a correct final answer can still yield high backward likelihood regardless of the reasoning path that produced it. Tighter integration with reasoning models---for example, combining instruct and reasoning models, or generating pseudo-reasoning for the forward pass---is an exciting direction for future work.
\section{LLM Usage}
LLM is used to polish writing in Section~\ref{sec:intro} and Section~\ref{sec:method}. LLM is used to improve the delivery of Appendix~\ref{app:format_reward}, \ref{app:limitation}. All generated writing is verified and fact-checked. 
\end{document}